\PassOptionsToPackage{table}{xcolor}
\documentclass[10pt,letterpaper]{article}

\usepackage[T1]{fontenc}
\usepackage{refrain_arxiv}

\usepackage{amsmath,amsfonts,bm}

\def\eqref#1{equation~\ref{#1}}

\def\1{\bm{1}}

\DeclareMathAlphabet{\mathsfit}{\encodingdefault}{\sfdefault}{m}{sl}
\SetMathAlphabet{\mathsfit}{bold}{\encodingdefault}{\sfdefault}{bx}{n}

\usepackage{amsmath}
\usepackage{amssymb}
\usepackage{booktabs}
\usepackage{graphicx}
\usepackage{wrapfig}
\usepackage{placeins}
\usepackage{needspace}
\usepackage{enumitem}
\usepackage{microtype}
\usepackage{multirow}
\usepackage{tabularx}
\usepackage{xcolor}
\usepackage{colortbl}
\usepackage{pifont}
\usepackage{fontawesome5}
\usepackage{hyperref}
\hypersetup{colorlinks=true,linkcolor=hunyuanblue,citecolor=hunyuanblue,urlcolor=hunyuanblue,
  pdftitle={Refrain: Learning Multi-Shot Consistency from Inconsistent References for A2O Video Generation},
  pdfauthor={Hongru Yan, Yue Wu, Jian-Wei Zhang, Liangbin Xie, Mingzhe Zheng, Shenghua Gao}}
\usepackage{url}

\newcommand{\refrain}{Refrain}
\newcommand{\durmix}{DurMix-300K}

\title{Refrain: Learning Multi-Shot Consistency from Inconsistent References for $A^2O$ Video Generation}

\author{
Hongru Yan\textsuperscript{1,2,*} \quad Yue Wu\textsuperscript{2,\ensuremath{\ddagger}} \quad Jian-Wei Zhang\textsuperscript{2} \quad Liangbin Xie\textsuperscript{2,3} \quad Mingzhe Zheng\textsuperscript{2,4} \quad Shenghua Gao\textsuperscript{1,\ensuremath{\dagger}}\\[7pt]
{\normalfont\fontsize{9}{11}\selectfont
\textsuperscript{*}Work done during internship at Tencent Hunyuan\\[2pt]
\textsuperscript{\ensuremath{\ddagger}}Project leader\qquad
\textsuperscript{\ensuremath{\dagger}}Corresponding author}\\[6pt]
{\fontsize{9}{11}\selectfont
\textsuperscript{1}The University of Hong Kong\quad
\textsuperscript{2}Tencent\quad
\textsuperscript{3}University of Macau\\[2pt]
\textsuperscript{4}The Hong Kong University of Science and Technology}
}

\begin{document}

\maketitle

\begin{abstract}
A comic-book character, a photorealistic cityscape, an oil-painted coffee cup --- such stylistically diverse references provide visual guidance across shots and frames, yet
generating a coherent multi-shot video that unifies their visual styles while preserving subject identities and scene content remains challenging.
Existing methods maintain cross-shot continuity through additional attention control or memory processing. However, they struggle to resolve disparate per-shot styles or handle reference--target duration mismatches within a single framework.
We introduce \textbf{Refrain}, a unified model that learns cross-shot consistency through a self-supervised reconstruction task. Reference clips from the same original video are deliberately perturbed with distinct visual edits, and the model recovers the unedited source footage under a shared style prompt. Refrain learns to suppress conflicting stylistic artifacts and retain reference content and motion dynamics. The resulting pipeline yields \textbf{DurMix-300K}, which pairs independently edited references of varying durations with their original videos. Operating on the $A^2O$ principle (\textbf{Any Reference, Any Duration, One Timeline}), Refrain maps image and video references onto a shared temporal canvas without auxiliary conditioning modules. For evaluation, experiments show that standard CLIP or DINO metrics frequently overlook stylistic clashes or misinterpret intended narrative cuts (such as changes in subjects or scenes) as inconsistency. We propose \textbf{Shot-Debiased Consistency (SDC)}, which uses an extensive real-video bank to establish a baseline for style variation associated with content changes within shots. By discounting this baseline divergence, SDC evaluates cross-shot style shifts against real-world distributions, where smaller residual differences indicate higher stylistic continuity. SDC agrees with professional raters and preserves method rankings across calibration banks containing up to 86.9M real videos.
Extensive experiments demonstrate that Refrain achieves state-of-the-art performance in cross-shot consistency, while excelling at shot timing, reference adherence, and text alignment.
\par\vspace{9pt}
{\sffamily\small\noindent\makebox[1.5em][l]{\faGithub}\textbf{Code available at: }\href{https://github.com/Tencent-Hunyuan/Refrain}{\texttt{https://github.com/Tencent-Hunyuan/Refrain}}\par}
\end{abstract}

\section{Introduction}
\label{sec:intro}
\begin{figure}[t]
    \centering
    \includegraphics[width=\linewidth]{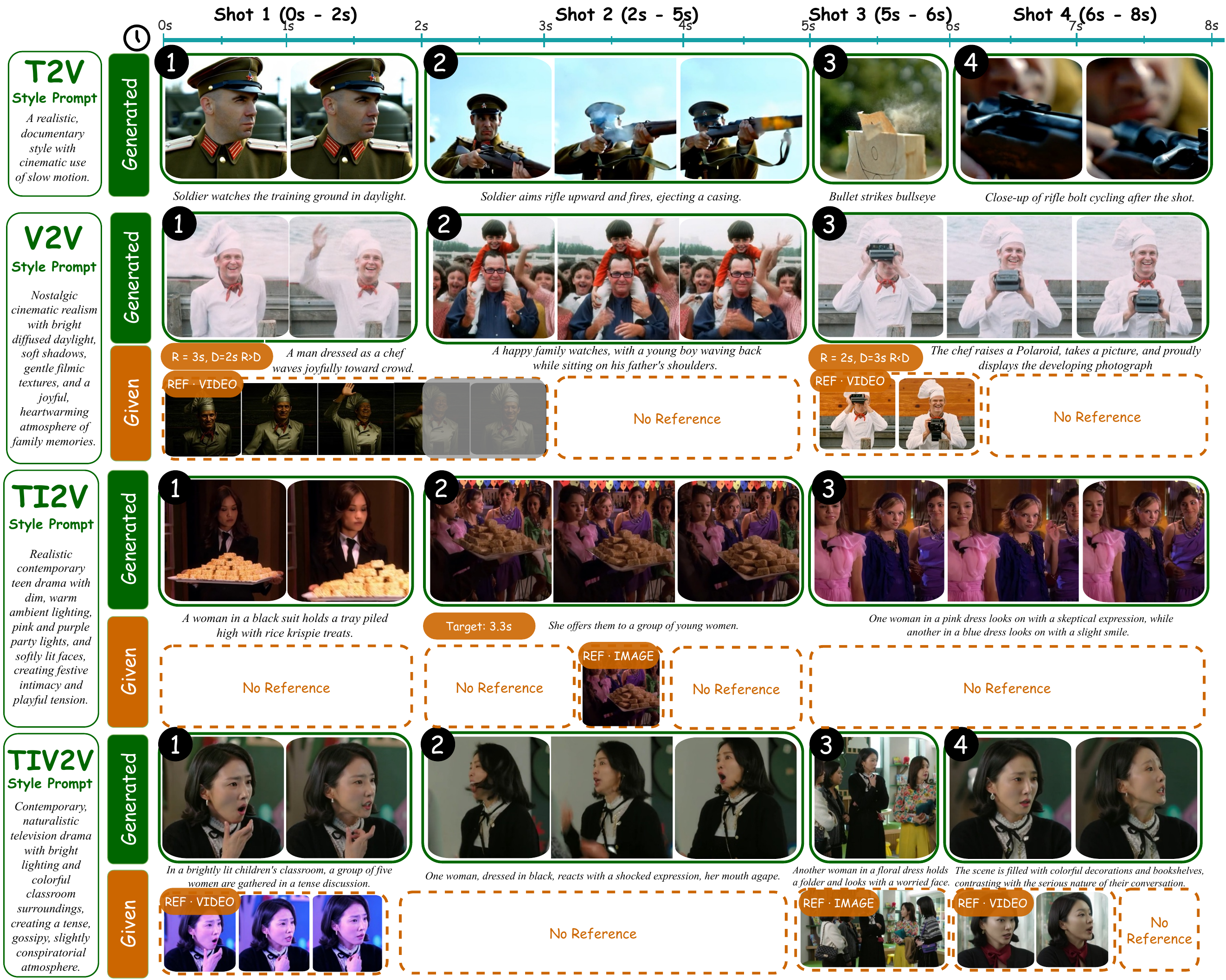}
    \caption{\textbf{Unified multi-shot generation with \refrain{}.} Each row shows a separate request with text, video, image, or joint conditioning. References vary in appearance and duration $R$, which can be shorter or longer than target duration $D$.}
    \label{fig:teaser}
\end{figure}

Generating coherent multi-shot videos from heterogeneous references (e.g., combining a photorealistic live-action clip with an animated reference) requires maintaining a unified visual style across shots while preserving underlying content and strict temporal boundaries (Figure~\ref{fig:teaser}). Existing frameworks use cross-reference attention, shot-specific masks, or visual memory to maintain continuity~\citep{liu2025phantom,meng2025holocine,zhang2025storymem}. Other methods support flexible condition placement or temporal control of reference subjects~\citep{cai2025videocanvas,girish2026alchemint}. We study the joint problem of resolving per-shot style conflicts and accommodating reference--target duration mismatches while preserving reference content and motion. References set the content of their shot intervals, and a shared style prompt sets the style (Section~\ref{sec:task}, Table~\ref{tab:task-compare}).

We propose Refrain, which learns cross-shot style consistency through a self-supervised reconstruction task built from ordinary videos (Figure~\ref{fig:method}). The name Refrain evokes a recurring musical phrase, reflecting visual continuity across shot transitions, while its ``Ref'' prefix highlights reference-conditioned generation. Refrain deliberately perturbs reference clips sampled from distinct shots of the same video with independent visual edits (e.g., style transfer, lighting variations, or weather shifts), introducing synthetic per-shot style conflicts. The model is then tasked with reconstructing the original target video under a shared style prompt. The generator receives the original shot descriptions and shared style prompt, without the edit instructions. Reconstructing the coherent target from independently edited references provides supervision for cross-shot style consistency. To support this supervision, we construct DurMix-300K, a large-scale dataset featuring over 300,000 reference conditions that pairs independently edited reference clips of arbitrary lengths with coherent target videos. 

Refrain jointly synthesizes the entire multi-shot sequence in a single diffusion trajectory, using shot-aware temporal position offsets to maintain sharp cut transitions. To accommodate heterogeneous multimodal references on this unified timeline, Refrain operates on the $A^2O$ principle: \textbf{Any Reference, Any Duration, One Timeline}. Specifically, reference images and variable-length video clips are mapped directly onto a shared temporal canvas, sparsely occupying their designated shot intervals via an occupancy mask without interpolation. By consolidating diverse inputs into a single sequence, a single backbone unifies text-to-video (T2V), video-to-video (V2V), text-image-to-video (TI2V), and composite (TIV2V) generation without auxiliary conditioning modules.

When evaluating cross-shot consistency of multi-shot videos, existing metrics like CLIP or DINO conflate legitimate narrative transitions with stylistic drift. They can overlook style conflicts and mistake content-feature changes caused by switching subjects or scenes for style inconsistency. We address this through a central question: \textit{how much style variation is naturally expected when content changes across shots?} To answer this, we propose Shot-Debiased Consistency (SDC), a metric that calibrates the natural coupling between content shifts and style fluctuations using a real-video (\textasciitilde{}2 million video clips) bank. SDC estimates the style variation associated with cross-shot content changes, subtracts this baseline, and compares the residual against real-video samples with comparable content changes.

Experiments show that independent-edit training improves consistency under per-shot inconsistent
test references while retaining reference adherence.
\refrain{} achieves state-of-the-art overall cross-shot consistency, shot timing, text alignment,
and reference adherence on the combined four-task benchmark. Our contributions:
\begin{itemize}[leftmargin=1.2em,itemsep=1pt,topsep=2pt]
    \item \textbf{Self-Supervised Data Construction and \durmix{}.} References from different shots are edited independently and cropped or extended to new durations, and the model learns to reconstruct the original video. Because the edited references conflict across shots while the target remains coherent, reconstruction supervises cross-shot style consistency without manual labels. From ordinary videos, this pipeline builds \durmix{} with over 300,000 reference conditions under 61 edit instructions in eight categories.
    \item \textbf{$A^2O$ Unified Generation.} One backbone supports T2V, V2V, TI2V, and TIV2V without added modules by mapping variable-duration references onto a shared timeline. Sparse placement keeps the reference latents without temporal interpolation, and shot-aware temporal coordinates mark the requested cuts. All shots are generated jointly in a single pass.
    \item \textbf{Shot-Debiased Consistency and Extensive Validation.} SDC measures cross-shot style consistency under intended content changes. It compares the style shift between shots with the shift observed in real videos at the same level of content change, which separates style drift from intended changes of subject or scene. Method rankings are identical across calibration banks of up to 86.9M real videos, and the scores agree with professional raters.
\end{itemize}

\begin{table}[t]
\centering\footnotesize
\caption{\textbf{Comparison of task settings.} Multi-shot generation~\citep{meng2025holocine,wang2025multishotmaster}, subject-to-video~\citep{liu2025phantom,wang2026refalign}, memory~\citep{zhang2025storymem}, and temporal placement~\citep{jiang2025vace,cai2025videocanvas}. \ding{51}: required; \ding{55}: not addressed.}
\label{tab:task-compare}
\setlength{\tabcolsep}{3pt}
\begin{tabularx}{\linewidth}{@{}>{\raggedright\arraybackslash}X*{5}{>{\centering\arraybackslash}p{1.95cm}}@{}}
\toprule
Setting & Refs & Duration $\neq D$ & Conflicting styles & Shared style & Cut timing \\
\midrule
Multi-shot & none/images & -- & \ding{55} & \ding{51} & \ding{51} \\
Subject-to-video & images & \ding{55} & \ding{55} & \ding{55} & \ding{55} \\
Memory multi-shot & images & \ding{55} & \ding{55} & \ding{51} & \ding{55} \\
Temporal placement & img.+video & \ding{51} & \ding{55} & \ding{55} & \ding{55} \\
\textbf{Ours} & img.+video & \ding{51} & \ding{51} & \ding{51} & \ding{51} \\
\bottomrule
\end{tabularx}
\end{table}

\section{Related Work}
\label{sec:related}

\paragraph{Multi-Shot and Reference-Conditioned Video Generation.}
Existing multi-shot and reference-guided generation frameworks typically maintain cross-shot continuity using auxiliary attention control, memory banks, or shot-specific masks~\citep{deng2025magref,girish2026alchemint,meng2025holocine,zhang2025storymem}. Other methods support flexible visual conditioning and reference placement~\citep{cai2025videocanvas,chen2026vino,guo2024sparsectrl,jiang2025vace}. Several works also build references from the target itself. Paint by Example and Video Alchemist augment the extracted reference before reconstruction~\citep{yang2023paintbyexample,chen2025videoalchemist}, Aura edits the background, lighting, clothing, or pose of subject references from single-shot clips~\citep{hu2026aura}, and MuSS takes subject references from other shots to avoid copying pose and viewpoint~\citep{li2026muss}. Refrain gives references of different shots independent, mutually conflicting appearance edits, supervises them with the coherent multi-shot source, and places image and video references of any duration on one timeline under the $A^2O$ principle, without auxiliary conditioning networks.

\paragraph{Evaluating Cross-Shot Consistency.}
Standard video benchmarks evaluate visual quality using general perceptual metrics or local feature similarities across cuts~\citep{huang2024vbench,zheng2025videogenofthought}. While Content-Debiased FVD addresses semantic bias in overall video distributions~\citep{ge2024cdfvd} and CSD targets visual style representations~\citep{somepalli2024csd}, existing metrics conflate legitimate content transitions (such as changes in subjects, scenes, or viewpoints) with genuine stylistic inconsistency. They fail to establish how much style variation is naturally expected across narrative cuts. Shot-Debiased Consistency (SDC) instead decouples style drift from semantic progression by benchmarking cross-shot style separation against within-shot dynamics calibrated on an extensive real-video bank.
\section{Method}
\label{sec:system}
\label{sec:unification}
\begin{figure}[t]
    \centering
    \includegraphics[width=\linewidth]{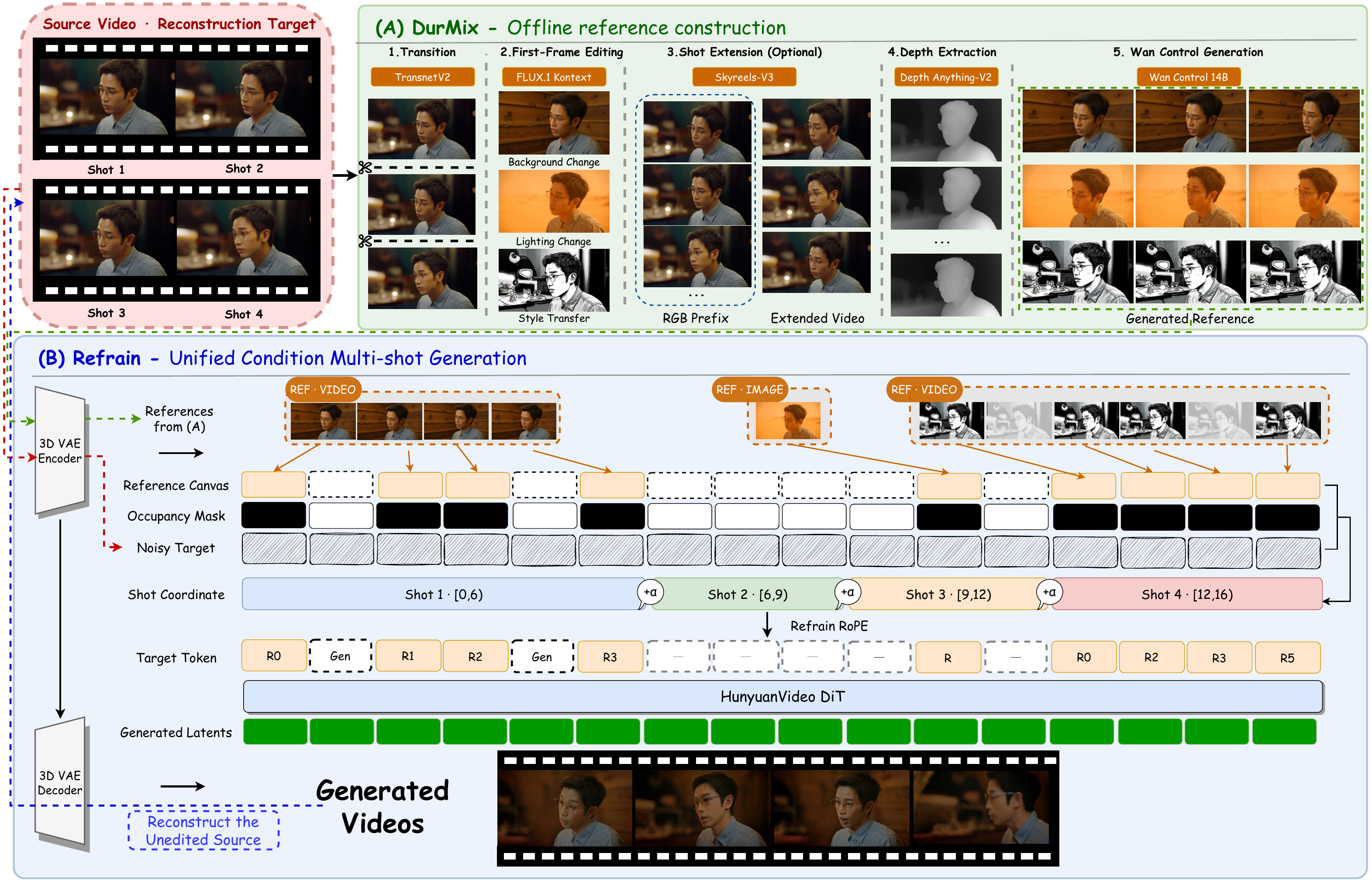}
    \caption{\textbf{Self-supervised reconstruction and unified conditioning.} (A) Reconstructing coherent source videos from independently edited references. (B) Joint multi-shot generation with sparse references on one timeline. Black indicates occupied mask positions ($m=1$).}
    \label{fig:method}
\end{figure}
Refrain learns cross-shot consistency through a self-supervised reconstruction task (Section~\ref{sec:data}). Independently edited references and their coherent source targets provide supervision for resolving per-shot style conflicts. The $A^2O$ canvas (Section~\ref{sec:setup-canvas}) represents image and video references of varying durations on the same target timeline, allowing one backbone to use this supervision across all four conditioning tasks (Figure~\ref{fig:method}).

To measure cross-shot visual style consistency without confounding content transitions or within-shot dynamics, we introduce Shot-Debiased Consistency (SDC) (Section~\ref{sec:sdc-method}).

\textbf{Task Definition.}\label{sec:task} A request specifies $K$ shots, a shared style prompt $s$, and $M\ge0$ image or video references. Shot $k$ has a description $y_k$ and spans $[t_k,t_{k+1})$. Reference $R_i$ is placed at $[d_i,d_i{+}D_i)$ on the same timeline, and its length $|R_i|$ can be shorter or longer than $D_i$. Reference appearances may conflict with the style specified by $s$ or with one another. The output should (i) render every shot in the style $s$, (ii) keep the content of each $R_i$, namely its subjects, layout, and motion, within $[d_i,d_i{+}D_i)$, and (iii) cut at the shot boundaries $t_k$. Table~\ref{tab:task-compare} compares this task with existing settings.

\subsection{Learning Cross-Shot Consistency through Self-Supervised Reconstruction}
\label{sec:data}
\label{sec:shortdata}

\textbf{Reconstruction Target and Reference Edits.} Let $V$ be a coherent source video, $y$ its original shot descriptions and shared style prompt, and $\mathcal R=\{R_i\}_{i=1}^{M}$ the selected references. An editing pipeline $\mathcal E$ transforms each reference using an instruction $e_i$ from a distribution $p_{\rm edit}$ over the instruction library. Before quality filtering, shared and independent editing differ in how these instructions are assigned:
\begin{equation}
 \widetilde R_i=\mathcal E(R_i;e_i),\qquad
 \begin{cases}
 e_1=\cdots=e_M=e,\quad e\sim p_{\rm edit},&\text{shared},\\
 e_1,\ldots,e_M\overset{\rm iid}{\sim}p_{\rm edit},&\text{independent}.
 \end{cases}
 \label{eq:edit-assignment}
\end{equation}
Independent draws can assign, for example, a style change to one reference and a lighting change to another. Both strategies pair the edited references $\widetilde{\mathcal R}$ with the unchanged target $V$. Reconstructing that target encourages the model to use useful reference content while following the requested output style. The generator receives $y$ and $\widetilde{\mathcal R}$ without the edit instructions.

\textbf{Data Construction Pipeline.} For each selected reference shot, FLUX.1 Kontext~\citep{labs2025fluxkontext} edits the first frame under one of 61 instructions in eight categories, drawn per Equation~\ref{eq:edit-assignment}. Depth-conditioned Wan-2.1~\citep{wanteam2025wan}, guided by Depth Anything v2~\citep{yang2024depthanythingv2}, then propagates the edit through the clip while keeping its motion. Shorter references are cropped in latent space. For references requiring extension, SkyReels-V3~\citep{li2026skyreelsv3} first completes the RGB control, and depth is then re-estimated before Wan propagation. Finally, Qwen3-VL-235B~\citep{bai2025qwen3vl} and TransNetV2~\citep{soucek2020transnetv2} remove clips with structural degradation or unintended cuts. Some edits also alter subject attributes, layout, or motion. For these edits, the unedited target still defines the content, and the model learns to recover it from partially altered references. At inference, a reference carries the intended content and is kept as given (Appendix~\ref{app:data}).

\paragraph{Training objective.}
We encode the original video as $z_0=\operatorname{Enc}(V)$ and form the standard flow-matching path from the clean latent to Gaussian noise~\citep{lipman2023flow}:
\begin{equation}
 x_\tau=(1-\tau)z_0+\tau\epsilon,\qquad
 u_\tau=\frac{\partial x_\tau}{\partial\tau}=\epsilon-z_0,
 \quad\epsilon\sim\mathcal N(0,I).
 \label{eq:flow-path}
\end{equation}
Here $\tau\in[0,1]$ is sampled from the training noise-level distribution $p_\tau$. Let $\mathcal D_{\rm ind}$ contain the retained independent-edit training pairs, and let $\Phi(\widetilde{\mathcal R})$ be their canvas representation from Section~\ref{sec:canvas}. The model learns to predict the target velocity in Equation~\ref{eq:flow-path}:
\begin{equation}
 \mathcal L_{\rm ind}(\theta)=
 \mathbb E_{\substack{(z_0,y,\widetilde{\mathcal R})\sim\mathcal D_{\rm ind}\\
 \tau\sim p_\tau,\;\epsilon\sim\mathcal N(0,I)}}
 \left\|v_\theta\!\left(x_\tau,\tau,y,\Phi(\widetilde{\mathcal R})\right)-u_\tau\right\|_2^2.
 \label{eq:ind-objective}
\end{equation}
Equation~\ref{eq:ind-objective} supervises all target locations, including those conditioned on edited references. Shared-edit training replaces $\mathcal D_{\rm ind}$ with $\mathcal D_{\rm shared}$ and retains the same loss. Both strategies use the reconstruction objective without an auxiliary consistency loss. Section~\ref{sec:editing-results} tests the two strategies empirically.

\subsection{Any Reference, Any Duration, One Timeline ($A^2O$)}
\label{sec:setup-canvas}
\label{sec:canvas}
\label{sec:sparse}
\label{sec:softref}
We use a shared reference canvas to support Text-to-Video (T2V), Video-to-Video (V2V), Text-Image-to-Video (TI2V), and Text-Image-Video-to-Video (TIV2V) in one backbone without task-specific adapter modules or auxiliary networks. Given a target latent space $z_0 \in \mathbb{R}^{C \times T \times H \times W}$, a zero-initialized canvas latent $c \in \mathbb{R}^{C \times T \times H \times W}$ and a binary conditioning mask $m \in \{0, 1\}^T$ track occupied temporal coordinates.
We concatenate the canvas and mask with the noisy target latents along channels to keep the token sequence length fixed. We compare this choice with sequence concatenation in Section~\ref{sec:free} (Table~\ref{tab:injection}(c)).

\textbf{Sparse Mapping}.
We use sparse assignment to accommodate reference--target duration mismatches.
In latent-frame coordinates, let $R$ be the reference length and $[d,d+D)$ its assigned interval. We map its latents $r_0, \dots, r_{R-1}$ to destination indices via exact assignment:
\begin{equation}
\label{eq:sparseexact}
\left\{
\begin{array}{lll}
\displaystyle
\pi(n) = d + \left\lfloor
\frac{n(D-1)}{R-1} + \frac{1}{2}
\right\rfloor,
&
c_{\pi(n)} = r_n,\quad m_{\pi(n)} = 1
&
(0 \le n < R \le D,\ R > 1),
\\[8pt]
\displaystyle
\rho(j) = \left\lfloor
\frac{j(R-1)}{D-1} + \frac{1}{2}
\right\rfloor,
&
c_{d+j} = r_{\rho(j)},\quad m_{d+j} = 1
&
(0 \le j < D < R,\ D > 1).
\end{array}
\right.
\end{equation}
For $1<R\le D$, reference latents are assigned without temporal interpolation, leaving unoccupied positions at $m=0$. For $R>D>1$, source latents are subsampled with both endpoints retained. If $R=1$ or $D=1$, only $r_0$ is placed at $c_d$, with $m_d=1$.

\textbf{Shot-Aware Coordinates.}
We adopt the shot-aware temporal offsets of MultiShotMaster~\citep{wang2025multishotmaster} in Rotary Position Embeddings (RoPE)~\citep{su2024roformer} to mark shot boundaries while retaining temporal self-attention across shots.
For a frame at temporal index $t$ belonging to shot $k(t) \in \mathbb{N}_0$, its adjusted coordinate $p(t)$ is:
\begin{equation}
    p(t) = t + \alpha k(t), \quad \alpha = 0.5.
\end{equation}

\begin{figure}[t]
    \centering
    \includegraphics[width=\textwidth]{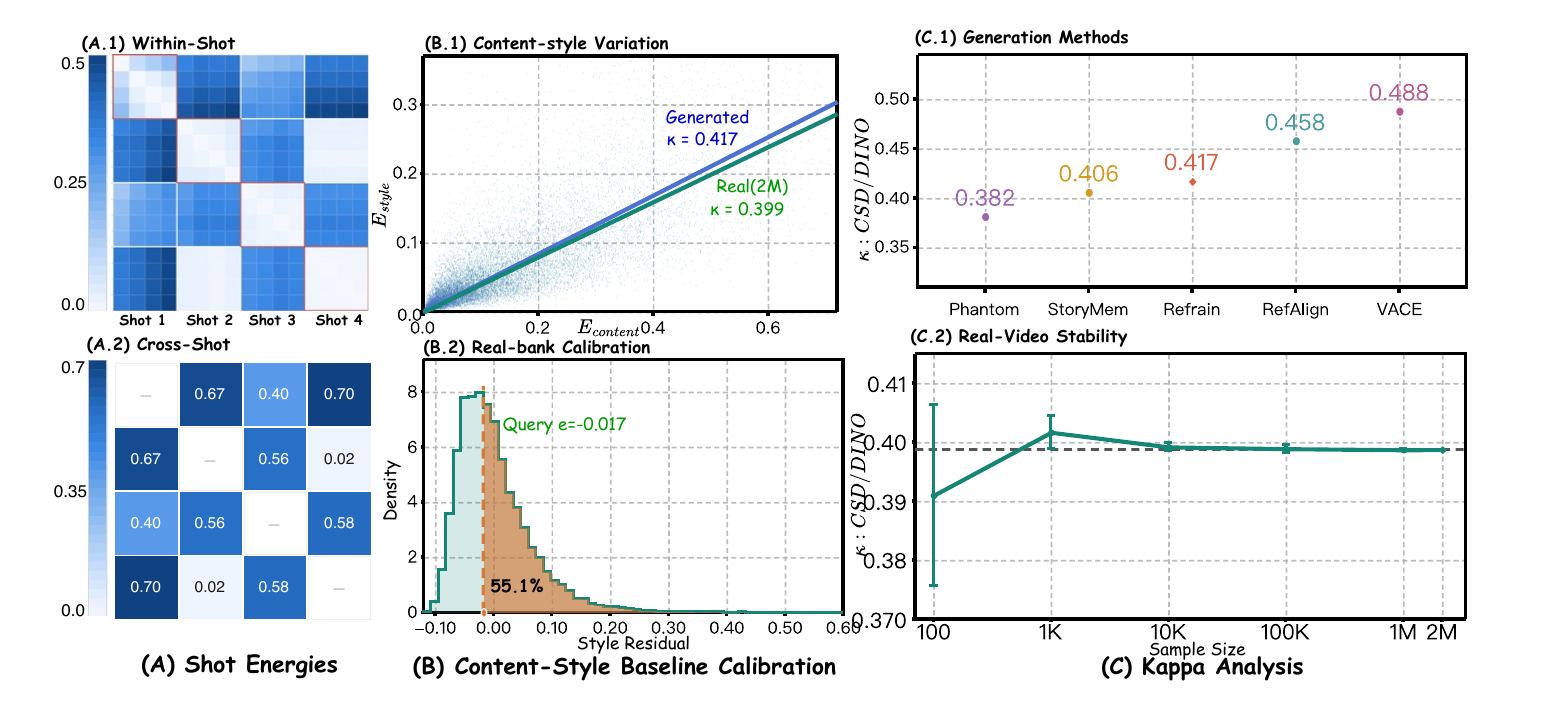}
    \caption{\textbf{SDC construction and motivation.} (A) Within- and cross-shot energies. (B) Content--style calibration. (C) Source-dependent $\kappa$ and sample-size stability.}
    \label{fig:sdc-internals}
\end{figure}

\subsection{Shot-Debiased Consistency}
\label{sec:sdc-method}

\textbf{Cross-Shot Energy Distance.} Evaluating raw frame features conflates cross-shot style drift with within-shot camera or object motion. We account for within-shot variation using an empirical energy contrast equivalent to squared maximum mean discrepancy (MMD) with a linear kernel~\citep{szekely2013energy,gretton2012mmd,sejdinovic2013equivalence}. Let $x_{i, a}^r$ be the normalized feature of frame $a$ in shot $i$. We extract style features (CSD~\citep{somepalli2024csd} and VGG AdaIN~\citep{huang2017adain}, $r \in \{\text{CSD}, \text{VGG}\}$) and content features (DINOv2 CLS~\citep{oquab2023dinov2}, $r = c$). Defining frame distance $d^r(x, y) = 1 - \langle x, y \rangle$ and shot mean $\mu_i^r = \frac{1}{n_i} \sum_{a=1}^{n_i} x_{i, a}^r$, the empirical cross-shot energy contrast $E_{ij}^r$ between shot $i$ and shot $j$ is (Figure~\ref{fig:sdc-internals}(A)):
\begin{equation}
\label{eq:energy}
    E_{ij}^r = 2 \cdot \underset{a, b}{\text{mean}} \, d^r(x_{i,a}^r, x_{j,b}^r) - \underset{a, a'}{\text{mean}} \, d^r(x_{i,a}^r, x_{i,a'}^r) - \underset{b, b'}{\text{mean}} \, d^r(x_{j,b}^r, x_{j,b'}^r) = \Vert{}\mu_i^r - \mu_j^r\Vert{}_2^2.
\end{equation}

\begin{figure}[!t]
    \centering
    \includegraphics[width=\textwidth]{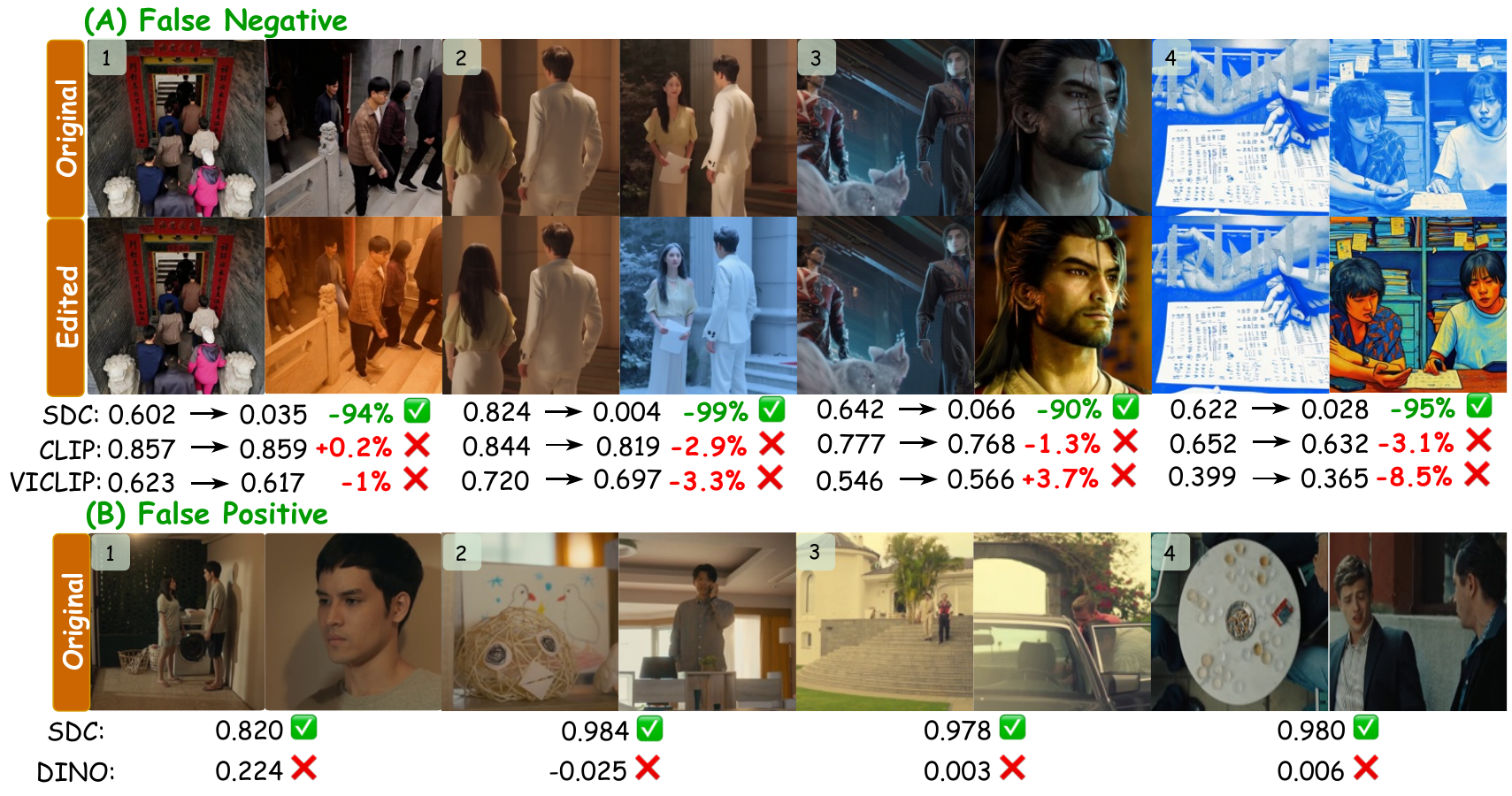}
    \caption{\textbf{Qualitative validation of SDC.} (A) SDC detects style edits missed by CLIP/ViCLIP. (B) SDC remains high across coherent scene cuts that lower DINO similarity.}
    \label{fig:sdc-compare}
\end{figure}

\textbf{Content-Style Baseline Calibration.} Frames within the same shot provide samples of natural variation without a shot transition. SDC uses these samples to establish a baseline for the style variation associated with content changes, then measures cross-shot style differences relative to this baseline. We construct a calibration bank from 2,144,201 real videos containing $6,462,692$ individual shots. Each shot is split into balanced frame sub-groups $A$ and $B$, yielding paired content and style split energies $C_b = \Vert{}\mu_A^c - \mu_B^c\Vert{}_2^2$ and $S_b^r = \Vert{}\mu_A^r - \mu_B^r\Vert{}_2^2$. We fit the expected content-to-style scaling slope $\hat{\kappa}_r$ via a through-origin Huber regression (Figure~\ref{fig:sdc-internals}(B.1)):
\begin{equation}
    \hat{\kappa}_r = \frac{\sum_b w_b a_b C_b S_b^r}{\sum_b w_b a_b C_b^2}, \quad a_b = \min\left(1, \frac{1.345 \hat{\sigma}}{\vert{}S_b^r - \hat{\kappa}_r C_b\vert{}}\right),
\end{equation}
where weights $w_b$ balance video, shot, and split frequencies, and $\hat{\sigma}$ is updated via the weighted median absolute deviation of residuals. The fitted slope $\hat{\kappa}_r$ provides an empirical content--style calibration for the chosen real-video bank and feature representations, with stable estimates across sample sizes (Figure~\ref{fig:sdc-internals}(C.2), Appendix~\ref{app:kappa}).

\textbf{Conditional Residual Ranking.} A video is scored by how much its style drifts beyond the expected baseline, compared against real videos with comparable content changes. For a query pair $(i, j)$ with energies $E_{ij}^c$ and $E_{ij}^r$, we subtract the expected content-induced shift to obtain the style residual:
\begin{equation}
    e_{ij}^r = E_{ij}^r - \hat{\kappa}_r E_{ij}^c
\end{equation}
Real-video residuals $z_b^r = S_b^r - \hat{\kappa}_r C_b$ are partitioned into 256 equal-mass content energy bins. The linear correction reduces content-related variation within each bin. For a query pair whose content energy $E_{ij}^c$ falls into bin $B$, its style rank is computed via the weighted upper-tail survival percentile (Figure~\ref{fig:sdc-internals}(B.2)):

\begin{equation}
    \hat{G}_{B, r}(e) = \frac{\sum_{b \in B} w_b \left[ \mathbb{I}(z_b^r > e) + \frac{1}{2} \mathbb{I}(z_b^r = e) \right]}{\sum_{b \in B} w_b}.
\end{equation}
Linear interpolation between bin centers in content-percentile space gives the conditional survival estimate $\hat{G}_r(e \mid E_{ij}^c)$. The query rank is:
\begin{equation}
    R_{ij}^r = \hat{G}_r(e_{ij}^r \mid E_{ij}^c) = \hat{G}_r \left( E_{ij}^r - \hat{\kappa}_r E_{ij}^c \;\middle\vert{}\; E_{ij}^c \right)
\end{equation}

Finally, for a video $V$ comprising $K \ge 2$ shots, we fuse the ranks from the CSD and VGG style heads in a fixed 2:1 ratio, and average over all $\binom{K}{2}$ shot pairs:

\begin{equation}
\label{eq:sdc-v5}
    \text{SDC}(V) = \frac{1}{\binom{K}{2}} \sum_{1 \le i < j \le K} \left(R_{ij}^{\text{CSD}}\right)^{2/3} \left(R_{ij}^{\text{VGG}}\right)^{1/3}. 
\end{equation}
At comparable content changes, smaller residual style differences receive higher real-bank ranks, yielding higher SDC scores and indicating greater cross-shot consistency.

\section{Experiments}
\label{sec:experiments}

\begin{figure}[!t]
    \centering
    \includegraphics[width=\linewidth]{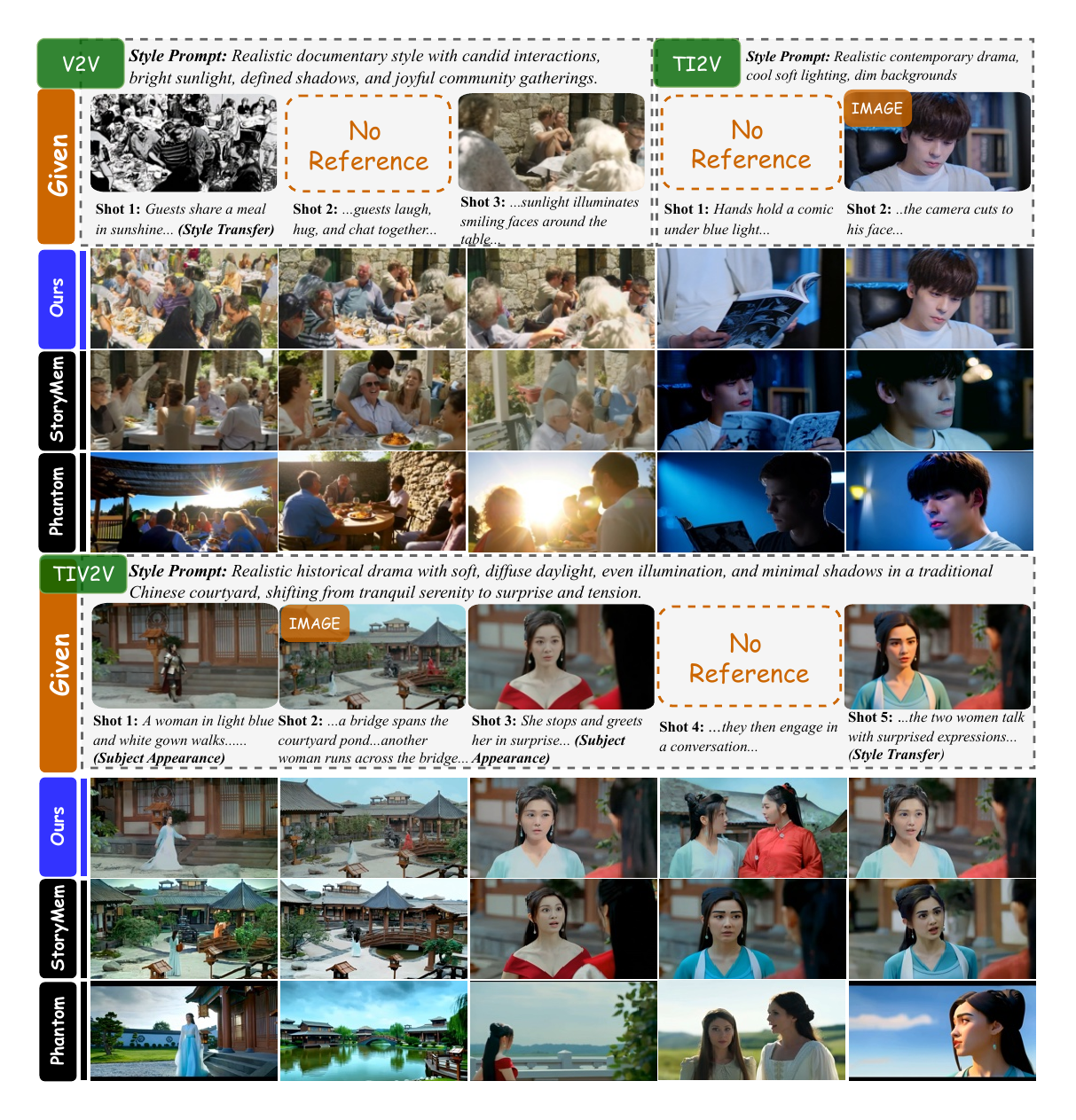}
    \caption{\textbf{Multi-shot generation from per-shot inconsistent references.}
    \refrain{} follows the reference content and shot descriptions while generating a shared
    visual style across shots.}
    \label{fig:qualitative-comparison}
\end{figure}

\subsection{Experimental Setup}
\label{sec:setup}

\textbf{Dataset and Task Allocation}. We collect 2,150,758 web videos and annotate their shots and visual styles with Gemini-3~\citep{pichai2025gemini3}
(example in Appendix~\ref{app:data}). We allocate 1,846,156 videos (85.84\%) to T2V
training, 297,192 (13.82\%) to reference editing, and 7,410 (0.34\%) to testing, and no source video is shared between training and test. We construct 300 evaluation requests from the test split across T2V, V2V, TI2V, and TIV2V. We construct DurMix-300K containing over 300,000 paired conditions with variable durations using the pipeline of Section~\ref{sec:data}. DurMix combines style edits with attribute perturbations. Style edits create per-shot appearance conflicts for learning cross-shot consistency, while changes to subject appearance, spatial relations, and motion provide additional perturbations for reconstruction. References are both shorter (200,265) and longer (109,087) than their targets.

Two independently collected sets with no overlap with our data serve only for evaluation: IP (17.2M clips of films, series, anime, and games) and Stock (67.6M everyday web videos such as tutorials and cooking). They extend the SDC calibration bank to 86.9M videos and supply cross-source footage for the human study.

\textbf{Implementation Details}. Refrain uses an 8.49B parameter HunyuanVideo-1.5 DiT architecture~\citep{wu2025hunyuanvideo} as its underlying backbone. The model is first pre-trained for 19,000 steps on pure T2V tasks using 256 NVIDIA H800 GPUs (around 5 days), followed by 26,000 reference fine-tuning steps using 16 H800 GPUs (around 2 days): 17,000 steps with references matched to target durations, then 9,000 steps with variable-duration references. Text-only tasks and examples without selected references maintain T2V training. Condition dropout (rate 0.2) retains at least one active reference.

\textbf{Baselines and Evaluation Metrics.} We compare Refrain with representative multi-shot and reference-conditioned generation methods, including StoryMem~\citep{zhang2025storymem}, Phantom~\citep{liu2025phantom}, RefAlign~\citep{wang2026refalign}, and VACE~\citep{jiang2025vace}. Cross-shot style consistency is evaluated using our proposed SDC. Transition timing accuracy is measured via Transition Mean Absolute Error (MAE). Text alignment is scored using CLIP-T and VideoAlign TA~\citep{liu2025videoalign}. Inter-shot consistency is assessed across semantics, subjects, and scenes. Reference fidelity is measured by semantic, subject, and background consistency.

\subsection{Results}
\label{sec:main}

Refrain leads the baselines in visual consistency, reference adherence, and temporal precision (Table~\ref{tab:fused300-main}).

\textbf{Style Consistency and Reference Fidelity.} Under test references with deliberate style conflicts, Refrain achieves the highest style consistency among the compared methods (SDC 0.315). Reference Semantic reaches 0.737, alongside leading reference subject (0.671) and background (0.674) scores (Figure~\ref{fig:qualitative-comparison}).

\textbf{Transition Accuracy and Text Alignment.} Refrain jointly generates all shots in a single diffusion trajectory and achieves a transition MAE of 0.199s. Refrain also has the highest VideoAlign TA (-0.221) and CLIP-T (0.222).

\begin{table*}[t]
\caption{\textbf{Unified multi-shot generation.} Bold/underlined: best/second-best; \textcolor{gray}{\ding{55}}: unreported or inapplicable. Parameters include all experts and unmerged adapters. FLOPs are analytical denoising estimates for a $480\!\times\!832$, 241-frame, three-shot video with three 49-frame references under each method's sampling schedule and reference adapter, excluding encoders and VAE decoding. SDC uses a fixed real-video bank. Metric definitions: Appendix~\ref{app:long-consistency}.}
\label{tab:fused300-main}
\centering\fontsize{7.5}{9}\selectfont
\setlength{\tabcolsep}{1.3pt}
\begin{tabularx}{\textwidth}{@{}>{\raggedright\arraybackslash}X*{12}{r}@{}}
\toprule
\multirow{2}{*}{Method} & \multirow{2}{*}{\shortstack{\#Params\\(B)}} & \multirow{2}{*}{\shortstack{FLOPs\\(P) $\downarrow$}} & \multicolumn{4}{c}{Generation and Alignment} & \multicolumn{3}{c}{Inter-Shot $\uparrow$} & \multicolumn{3}{c}{Reference $\uparrow$} \\
\cmidrule(lr){4-7}\cmidrule(lr){8-10}\cmidrule(lr){11-13}
& & & MAE\,s $\downarrow$ & CLIP-T $\uparrow$ & TA $\uparrow$ & SDC $\uparrow$ & Semantic & Subject & Scene & Semantic & Subject & Background \\
\midrule
HunyuanVideo1.5 & 8.49 & \textbf{324.4} & \underline{0.802} & 0.212 & -0.386 & 0.090 & 0.566 & 0.439 & 0.478 & \textcolor{gray}{\ding{55}} & \textcolor{gray}{\ding{55}} & \textcolor{gray}{\ding{55}} \\
Phantom & 14.29 & 648.7 & \textcolor{gray}{\ding{55}} & 0.203 & -0.675 & 0.130 & 0.491 & 0.345 & 0.385 & 0.556 & 0.450 & 0.376 \\
RefAlign & 14.29 & 807.7 & \textcolor{gray}{\ding{55}} & 0.204 & -0.530 & 0.125 & 0.493 & 0.341 & 0.361 & \underline{0.614} & \underline{0.469} & \underline{0.448} \\
StoryMem & 29.80 & \underline{493.2} & \textcolor{gray}{\ding{55}} & 0.197 & -0.679 & \underline{0.175} & \underline{0.708} & \underline{0.489} & \textbf{0.531} & 0.478 & 0.412 & 0.388 \\
VACE & 17.34 & 608.9 & \textcolor{gray}{\ding{55}} & \underline{0.217} & \underline{-0.336} & 0.108 & 0.492 & 0.362 & 0.361 & \textcolor{gray}{\ding{55}} & \textcolor{gray}{\ding{55}} & \textcolor{gray}{\ding{55}} \\
\midrule
\rowcolor{gray!15}
\textbf{\refrain{} (ours)} & 8.49 & \textbf{324.4} & \textbf{0.199} & \textbf{0.222} & \textbf{-0.221} & \textbf{0.315} & \textbf{0.710} & \textbf{0.494} & \underline{0.491} & \textbf{0.737} & \textbf{0.671} & \textbf{0.674} \\
\bottomrule
\end{tabularx}
\end{table*}

\subsection{Ablation Studies}
\begin{table}[htbp]
\centering\scriptsize
\caption{\textbf{Ablation studies.}
(a) Reference editing. (b) Shot coordinates and placement. (c) Channel vs.\ sequence injection; Tok.: token count relative to the target.}
\label{tab:reference-edit}
\label{tab:fused300-2x2}
\label{tab:injection}
\setlength{\tabcolsep}{2pt}
\begin{minipage}[t]{0.29\linewidth}
\centering
\textbf{(a) Reference editing}\par\smallskip
\begin{tabular*}{\linewidth}{@{\extracolsep{\fill}}llr@{}}
\toprule
\textbf{Train edits} & \textbf{Test edits} & \textbf{SDC}$\uparrow$ \\
\midrule
Independent & Independent & \textbf{0.315} \\
Independent & Shared & 0.299 \\
Shared & Independent & 0.301 \\
Shared & Shared & 0.304 \\
\bottomrule
\end{tabular*}

\end{minipage}\hfill
\begin{minipage}[t]{0.33\linewidth}
\centering
\textbf{(b) Coordinates, placement}\par\smallskip
\begin{tabular*}{\linewidth}{@{\extracolsep{\fill}}llrr@{}}
\toprule
\textbf{Coordinates} & \textbf{Placement} & \textbf{MAE}\,s$\downarrow$ & \textbf{SDC}$\uparrow$ \\
\midrule
\textbf{Shot-aware} & \textbf{Sparse} & \textbf{0.199} & \textbf{0.315} \\
Shot-aware & Resample & 0.233 & 0.310 \\
None & Sparse & 0.281 & 0.304 \\
None & Resample & 0.293 & 0.289 \\
\bottomrule
\end{tabular*}

\end{minipage}\hfill
\begin{minipage}[t]{0.33\linewidth}
\centering
\textbf{(c) Injection}\par\smallskip
\begin{tabular*}{\linewidth}{@{\extracolsep{\fill}}lcccc@{}}
\toprule
\textbf{Variant} & \textbf{SDC}$\uparrow$ & \textbf{Ref-S}$\uparrow$ & \textbf{MAE}$\downarrow$ & \textbf{Tok.} \\
\midrule
Channel & \textbf{0.315} & \textbf{0.671} & \textbf{0.199} & 1.00$\times$ \\
Seq. & \textbf{0.315} & 0.666 & 0.209 & 1.64$\times$ \\
\bottomrule
\end{tabular*}

\end{minipage}
\end{table}

\label{sec:free}
\label{sec:q1}

\paragraph{Impact of Inconsistent Reference Supervision.}
\label{sec:editing-results}
We compare independent and shared reference-edit instructions under the same training budget (Equation~\ref{eq:edit-assignment}). On independently edited test references, independent-edit training raises SDC from 0.301 to 0.315. With shared-edit test references, independent-edit training reduces SDC from 0.304 to 0.299 (Table~\ref{tab:reference-edit}(a)). The gain occurs when the test references contain per-shot inconsistency.

\begin{wraptable}{r}{0.35\textwidth}
\centering\small
\vspace{-\intextsep}
\vspace{-\abovecaptionskip}
\caption{\textbf{Pure T2V generation.}}
\label{tab:t2v-transfer}
\setlength{\tabcolsep}{4pt}
\begin{tabular*}{\linewidth}{@{\extracolsep{\fill}}lr@{}}
\toprule
\textbf{Training}  & \textbf{SDC} $\uparrow$ \\
\midrule
T2V pretraining  & 0.253 \\
Independent edits & \textbf{0.303} \\
\bottomrule
\end{tabular*}
\end{wraptable}

We also evaluate the same text-only requests before and after independent-edit fine-tuning, without visual references at inference. SDC increases from 0.253 after T2V pretraining to 0.303 after fine-tuning (Table~\ref{tab:t2v-transfer}), showing a consistency gain beyond reference-conditioned generation.

\paragraph{Shot Coordinates and Reference Placement.}
Table~\ref{tab:fused300-2x2}(b) verifies the core components of the $A^2O$ canvas. Shot-aware RoPE offsets matter for boundary precision. Removing them increases transition error from 0.199s to 0.281s. Replacing our sparse latent placement with temporal frame resampling worsens transition error to 0.233s while reducing SDC from 0.315 to 0.310. Sparse placement avoids latent interpolation. Appendix~\ref{app:duration-examples} shows how Refrain adapts reference actions to the requested shot duration while preserving natural motion pacing and a shared visual style.

\paragraph{Channel vs.\ Sequence Injection.} Sequence concatenation reaches the same SDC (0.315) with lower TA ($-$0.250 vs.\ $-$0.221) and higher MAE (0.209 vs.\ 0.199\,s) at 1.8$\times$ the step time (Table~\ref{tab:injection}(c), Appendix~\ref{app:rebuttal-exps}).

\subsection{Validating SDC}
\begin{table}[!t]
\centering
\caption{\textbf{Style and content controls.}
(a) Single-shot style edits (108,909 videos). (b) Shared/independent edits (72,570 videos).
Means and detection at 5\% control false positives. (c) Real videos.}
\label{tab:sdc-population}
\begingroup
\footnotesize
\setlength{\tabcolsep}{2pt}
\begin{minipage}[t]{0.31\linewidth}
\centering
\textbf{(a) Single-shot style edits}\par\smallskip
\begin{tabularx}{\linewidth}{@{}X@{\hspace{4pt}}cc@{}}
\toprule
Metric & Score & Detect. $\uparrow$ \\
\midrule
CLIP & $0.836\!\to\!0.711$ & 48.6\% \\
ViCLIP & $0.609\!\to\!0.426$ & 38.6\% \\
DINO & $0.385\!\to\!0.297$ & 10.0\% \\
\rowcolor{gray!12} \textbf{SDC} & $0.346\!\to\!0.047$ & \textbf{78.0\%} \\
\bottomrule
\end{tabularx}
\end{minipage}\hfill
\begin{minipage}[t]{0.31\linewidth}
\centering
\textbf{(b) Independent reference edits}\par\smallskip
\begin{tabularx}{\linewidth}{@{}X@{\hspace{4pt}}cc@{}}
\toprule
Metric & Score & Detect. $\uparrow$ \\
\midrule
CLIP & $0.771\!\to\!0.704$ & 15.2\% \\
ViCLIP & $0.560\!\to\!0.420$ & 18.9\% \\
DINO & $0.408\!\to\!0.312$ & 10.8\% \\
\rowcolor{gray!12} \textbf{SDC} & $0.302\!\to\!0.078$ & \textbf{50.9\%} \\
\bottomrule
\end{tabularx}
\end{minipage}\hfill
\begin{minipage}[t]{0.34\linewidth}
\centering
\textbf{(c) Real-video content changes}\par\smallskip
\begin{tabularx}{\linewidth}{@{}X@{\hspace{4pt}}rr@{}}
\toprule
Subset & DINO & SDC \\
\midrule
All (2.14M) & 0.383 & 0.337 \\
$\mathrm{DINO}<0.25$ & 0.170 & 0.339 \\
Lowest DINO 10\% & 0.117 & 0.322 \\
$\mathrm{DINO}<0.10$ & 0.059 & 0.287 \\
\bottomrule
\end{tabularx}
\end{minipage}
\endgroup
\end{table}

\label{sec:sdc-validation}

\paragraph{Style and Content Changes.}
Table~\ref{tab:sdc-population} extends Figure~\ref{fig:sdc-compare} to large-scale controls.
At 5\% false positives, SDC detects 78.0\% of single-shot style edits, versus 48.6\%
for CLIP and 38.6\% for ViCLIP. It also better distinguishes independent from shared
reference edits when both groups are edited. Among 2.14M real videos, the low-DINO
subset ($<0.25$) retains mean SDC 0.339 versus 0.337 overall, and large content changes do not lower SDC. Appendix~\ref{app:sdc} compares SDC with CSD/VGG
similarity and tests $\kappa=0$ (Tables~\ref{tab:sdc-style-similarity}--\ref{tab:sdc-kappa-ablation}).

\paragraph{Calibration Stability and Source Variation.}
\label{sec:kappa-analysis}
The real-video content--style slope remains close to 0.399 across calibration sizes from 100 to 2,144,201 videos (Figure~\ref{fig:sdc-internals}(C.2)). Generated-source slopes range from 0.382 to 0.488 and overlap the real-video estimate (Figure~\ref{fig:sdc-internals}(C.1), Appendix~\ref{app:kappa}). With IP and Stock added, the 86.9M-video bank gives a slope of 0.411 and a mean real-video SDC of 0.326, close to 0.337 on the paper bank. The method ranking is unchanged (Tables~\ref{tab:kappa-scaling} and~\ref{tab:sdc-bank-scaling}).

\paragraph{Human Agreement.}
Six professional raters rated 150 clips: real two-shot splices from our sources and from the two external collections, natural multi-shot clips, and generated clips. Over the 140 scorable clips, SDC agrees with the mean rating at $\rho=0.68$ and separates consensus-consistent from consensus-inconsistent clips with AUC 0.90. On real footage it reaches AUC 0.97 (Table~\ref{tab:sdc-human-v2}, Appendix~\ref{app:human}).

\begin{table}[htbp]
\centering\small
\caption{\textbf{SDC and human style ratings.} Consensus labels require at least four of six votes. $\rho$: Spearman correlation with the mean rating.}
\label{tab:sdc-human-v2}
\setlength{\tabcolsep}{6pt}
\begin{tabular}{lccc}
\toprule
Clips & Scored & Consistent / inconsistent & $\rho$ / AUC \\
\midrule
\textbf{All} & \textbf{140} & \textbf{73 / 48} & \textbf{0.68 / 0.90} \\
Real footage & 90 & 40 / 36 & 0.77 / 0.97 \\
Real, external collections & 54 & 25 / 22 & 0.79 / 0.98 \\
\bottomrule
\end{tabular}
\end{table}

\section{Conclusion}
\label{sec:apps}
\label{sec:limitations}
\label{sec:conclusion}

\refrain{} learns cross-shot consistency  through the self-supervised reconstruction task on
\durmix{}. Its shared timeline unifies text, image, and video conditioning without added
modules. SDC evaluates the resulting consistency against natural content--style variation
in real videos. Experiments show improved consistency and reference adherence under
per-shot inconsistent inputs, with consistency gains extending to pure T2V generation.
\label{sec:main-text-end}

\subsection*{AI use statement}

Large language models assisted with writing and code development, including analysis scripts.
The authors are responsible for the manuscript and reported results. Gemini-3 provides
video annotations, and Qwen3-VL provides data-quality filtering
(Section~\ref{sec:data} and Appendix~\ref{app:data}).

\bibliography{references}
\bibliographystyle{refrain_preprint}

\appendix
\section{Data Construction}
\label{app:data}

\paragraph{Source videos and annotations.}
We collect 2,150,758 web videos and use Gemini-3 to annotate shot boundaries, shot
descriptions, entities, and visual style. The split assigns 1,846,156 videos (85.84\%)
to T2V training, 297,192 (13.82\%) to reference-editing construction, and 7,410
(0.34\%) to testing. The 300 evaluation requests are drawn from the test allocation.
Original-video annotations provide the shot descriptions and shared style prompt.
Table~\ref{tab:annotation-example} presents an example annotation.

\begin{table}[h]
\centering\small
\caption{Example Gemini-3 annotation.}
\label{tab:annotation-example}
\begin{tabularx}{\linewidth}{@{}lX@{}}
\toprule
Field & Annotation \\
\midrule
Style & Realistic, cinematic thriller style. \\
Environment & A dimly lit urban street at night, lined with parked cars and trees.
The setting suggests a residential or commercial area after hours. \\
Shot 1, 0.0--2.7\,s & A woman (\texttt{<ID:1>}) holding a handgun
(\texttt{<ID:2>}) looks over her shoulder and speaks. \\
Shot 2, 2.7--4.2\,s & A man walks away from the camera towards a motorcycle on a dark street. \\
\bottomrule
\end{tabularx}
\end{table}

\durmix{} pairs per-shot inconsistent references with coherent original videos to supervise
reconstruction with a shared output style. Its construction plan contains 508{,}041 reference conditions,
with 495{,}634 assigned to training and 12{,}407 to testing. The validated inventory contains
309{,}352 conditions: 200{,}265 shorter and 109{,}087 longer than their target intervals.
These conditions belong to 223,850 source videos. The dataset's 300K designation refers to
the reference-condition count. Among 76{,}532 videos with
multiple video references, 80.03\% use different edit categories and 98.49\% use different
instructions.

\paragraph{Editing and quality filtering.}
To vary reference appearance and scene attributes, we sample from 61 instructions in eight
categories (Table~\ref{tab:edit-categories}). Each selected frame is edited and propagated through
its reference clip as described in Section~\ref{sec:data}. Qwen3-VL-235B evaluates preservation of
the reference, edit quality, and temporal validity. TransNetV2 rejects transitions inside intended
single-shot controls. For motion and spatial-relation edits, filtering evaluates reconstruction
suitability while accommodating the intended attribute changes.

\begin{table}[h]
\centering\small
\caption{Editing categories in \durmix{}. Counts refer to validated reference conditions.}
\label{tab:edit-categories}
\begin{tabular}{lrr}
\toprule
Category & Instructions & Conditions \\
\midrule
Style transfer & 20 & 125{,}457 \\
Subject appearance & 8 & 31{,}655 \\
Object material or colour & 7 & 24{,}544 \\
Lighting and weather & 6 & 31{,}175 \\
Background retexturing & 6 & 28{,}718 \\
Spatial relation & 6 & 21{,}266 \\
Camera or lens filtering & 4 & 22{,}672 \\
Motion change & 4 & 23{,}865 \\
\midrule
Total & 61 & 309{,}352 \\
\bottomrule
\end{tabular}
\end{table}

\paragraph{Roles of reference edits.}
All edits modify the reference inputs while leaving the reconstruction target unchanged.
Style transfer, camera filtering, and lighting edits primarily introduce appearance differences
across shots. Subject-appearance, spatial-relation, and motion edits additionally perturb
attributes of the source content. Attribute perturbations train reconstruction from partially
altered references. Recovery of an edited attribute depends on the information available
in the remaining references and text. Reference preservation concerns the source
information retained after editing, not every attribute introduced by an edit.
For example, a clothing or action change in a training reference does not change the clothing
or action required by the target. During training, the original video and its annotations
define the desired output, and edited references provide additional visual information.
At inference, a reference carries the content the user intends, and the content requirement of Section~\ref{sec:task} applies to it as given.

\paragraph{Duration construction.}
Reference duration varies while the selected reference frame, control start, and generation
target remain fixed. To avoid including a nearby transition, we remove exactly one latent
step, or four pixel frames, from the retained real-control endpoint. Existing short conditions are cropped
directly in latent space and keep only frames driven by real control. Their endpoint is
\(\min(F_{\mathrm{planned}},F_{\mathrm{real}})-4\) pixel frames.
Here $F_{\mathrm{planned}}$ is the assigned reference length and $F_{\mathrm{real}}$ is the
available real-control length, both in pixel frames. For RGB extension, SkyReels-V3
fills the gap between the tail-trimmed real control and the assigned reference duration. The completed
sequence preserves the exact real prefix. We re-estimate depth over that sequence before Wan
Control propagation and latent encoding, retaining the completed sequence length. Extension
supplies footage needed to construct a reference while preserving the requested output duration.
The resulting reference can remain shorter or longer than its generation target.

\paragraph{Construction resources.}
Generation of \durmix{} consumed an estimated 126 hours on 864 H20 GPUs.
TransNetV2 filtering used 128 H20 GPUs for approximately 44 minutes.
Qwen3-VL-235B, deployed on 128 H20 GPUs, performed screening and rechecking in approximately 16.4 hours. It screens edit quality, content preservation, and temporal validity on three source and three edited frames, with six of each for rechecking rejected clips. Among videos with several video references, 80.03\% receive different edit categories.

\section{Training and Inference Settings}
\label{app:controls}
\label{app:switches}

The backbone has 24 double-stream blocks, hidden width 3,072, 24 attention heads, 48 latent
channels, and approximately 8.49B parameters. The temporal and spatial RoPE dimensions are
$[32,48,48]$. The conditioning projection expands 48 channels to 97,
adding $49\times3072=150{,}528$ weights. Reference slots retain unit loss weight. Full
bidirectional attention uses text-padding masking. Reference conditioning uses channel concatenation,
and training uses the reconstruction objective. Table~\ref{tab:switches}
summarizes the training and inference settings.

\begin{table}[h]
\centering\small
\caption{Main fine-tuning and inference configuration. Conditioning probabilities act at the
segment/anchor level, and the resulting task frequencies depend on reference availability and selection.}
\label{tab:switches}
\begin{tabular}{ll}
\toprule
Setting & Value \\
\midrule
Pure T2V, steps 0--19k & 19{,}000 updates, 256 H800 GPUs, LR $10^{-5}$ \\
Reference fine-tuning, steps 19k--36k & 17{,}000 updates, 16 H800 GPUs, LR $10^{-5}$ \\
Reference fine-tuning, steps 36k--45k & 9{,}000 updates, 16 H800 GPUs, LR $2\times10^{-6}$ \\
Global batch size, steps 0--19k & 128 videos per optimizer update \\
Global batch size, steps 19k--36k & 8 videos per optimizer update \\
Global batch size, steps 36k--45k & 8 videos per optimizer update \\
Training clip length & 241 frames \\
Spatial preset & 480p, aspect-ratio dependent \\
Optimizer & AdamW \\
Warmup, continuation & 200 updates \\
Configuration seed & 42 \\
Per-segment condition probability & 0.6 \\
Image-anchor probability & 0.8 \\
Condition dropout probability & 0.2 \\
Conditioning path & Sparse reference placement, channel concatenation \\
Shot-coordinate offset & $\alpha=0.5$ \\
Sequence-parallel size, final fine-tuning & 8 \\
Inference steps / CFG / flow shift & 50 / 6.0 / 7.0 \\
\bottomrule
\end{tabular}
\end{table}

\paragraph{Training data and task sampling.}
The first reference-training stage fits references to their target lengths for 17,000 steps. Its base task
sampler assigns probabilities 0.3 to text-only generation and 0.7 to reference conditioning.
Dataset references are sampled separately, with probability 0.6 per available video reference
and 0.8 for an image anchor. Selected dataset references activate conditioning regardless
of the base task draw. Condition dropout removes individual references with probability 0.2 while retaining at
least one when references are active.

The final 9,000 updates use all 297,192 source-video records with the filtered \durmix{}
conditions and retain 117,868 records from the earlier reference data, a 71.60:28.40 ratio
by record count. References retain their native lengths and are placed sparsely on the target
timeline. Dataset reference sampling remains active, while examples without selected
references use text-only training. Both the main results and the complete
coordinate-by-placement comparison use 45,000 training steps.

\section{Evaluation Sets and Metrics}
\label{app:benchmark}
\label{app:metrics}

The main benchmark evaluates unified generation on 300 requests: 30 T2V, 30 TI2V, 120 V2V,
and 120 TIV2V.

In each of V2V and TIV2V, 40 requests (one third) replace the original descriptions
of video-referenced shots with reference-control instructions. The remaining 80 requests
retain those descriptions and append the control instructions. The shared style prompt
and descriptions of shots without video references are retained, including image-anchor shots.
The instructions are:
\begin{itemize}
    \item \textbf{Video reference:} ``Use the provided reference video segment for this time
    range and preserve its subject identity, motion, camera movement, and temporal dynamics.''
    \item \textbf{Image reference:} ``Use the provided reference image anchor at latent frame
    $t$ and preserve the subject identity and appearance around that moment.''
\end{itemize}
Here $t$ specifies the image-anchor position. Image instructions are appended to the
corresponding shot descriptions. During evaluation, image anchors are processed before
video references. A reference whose destination interval overlaps an already accepted
reference is skipped with a warning.

\paragraph{Transition timing.}
We match TransNetV2-detected cuts to the requested boundaries using order-preserving
dynamic programming. Matched cuts contribute absolute timing error, and each missing
or extra cut adds a one-second penalty. The total timing error in seconds is normalized by the larger cut count.
Transition MAE averages the normalized timing error over videos.

\paragraph{Text alignment.}
CLIP-T averages CLIP~\citep{radford2021clip} similarity between generated frames and their corresponding shot
descriptions. TA scores each requested shot with the pretrained VideoAlign model and its
published normalization~\citep{liu2025videoalign}, yielding a signed reward.
TA uses the original shot description and shared style prompt, including for shots whose
inference descriptions were replaced by reference-control instructions. Scores are averaged
across shots and then across videos.

\paragraph{Content consistency.}
\label{app:long-consistency}
We adapt the consistency categories of MultiShotMaster~\citep{wang2025multishotmaster}
to requests that explicitly change subjects and scenes. For each request, a fixed annotation
of the original shot descriptions identifies recurring entities, shared scenes, and
semantically comparable shot pairs. These annotations are shared by all methods.
Inter-Shot Semantic averages ViCLIP~\citep{wang2024internvid} cosine similarity over the annotated pairs, using
eight uniformly sampled frames per requested shot. Inter-Shot Subject compares only the
same entity across shots, and Inter-Shot Scene compares visible backgrounds only for
shots annotated as sharing a scene.

GroundingDINO~\citep{liu2024groundingdino} and SAM~\citep{kirillov2023sam} locate and segment subjects. We compute DINOv2-small cosine
similarities on subject crops with non-subject pixels masked, or on full frames with
foreground regions masked for background comparisons. First, middle, and last frames
are sampled independently from each interval, and similarities average all available
keyframe pairs. Reference Subject and Reference Background use the corresponding
reference and destination shot. Images use their exact anchor frames. Deliberate
subject-appearance or material edits are excluded from reference-subject comparisons, and
background, layout, or material edits are excluded from reference-background comparisons.
Reference Background excludes 82 references with background, spatial-layout, or
material/color edits.

Reference Semantic measures ViCLIP cosine similarity between each video reference and
its corresponding generated shot. Eight frames are uniformly sampled from the full native
reference and, independently, from its requested destination interval. Each eight-frame
sequence is resized to $224\times224$, normalized, and encoded by ViCLIP. Cosine
similarities are averaged over references within each video and then across videos.
The evaluation covers all 489 video references of the V2V and TIV2V requests, including
all edit categories. Image anchors are excluded from this video-level metric.
Destination intervals follow the fixed benchmark timeline used for the other content
metrics, without adjustment to detected cuts or output length. Each reported method
has valid scores for all 489 pairs.

Each metric compares methods on the same shot
or reference pairs and weights videos equally. These feature similarities measure cross-shot content correspondence.
SDC uses detected shots as defined in Section~\ref{sec:sdc-method}.

\subsection{Baseline Configurations}
\label{app:baseline-configurations}

\paragraph{Generation protocol.}
Phantom, RefAlign, StoryMem, and VACE generate individual shots and concatenate them
at the requested boundaries. Each shot receives the shared style description and its
own conditioning text, and all shots within a request use the same random seed.
Concatenation uses hard cuts without overlapping frames or transition blending. For
models requiring a $4k+1$ frame count, the extra terminal frame of each non-final shot
is removed before concatenation. Since this procedure sets the shot boundaries
externally, Transition MAE is unreported for these four methods. HunyuanVideo1.5
and Refrain generate the complete sequence jointly and are evaluated for
transition timing.

\paragraph{HunyuanVideo1.5.}
We evaluate the pretrained backbone before our multi-shot and reference-conditioned
training. Shot-aware temporal offsets are disabled. Image and video references are
sparsely placed in the existing conditioning channels, together with the occupancy
mask. The output latents remain generated throughout denoising, including at reference
locations. Sampling uses 50 steps, guidance scale 6.0, flow shift 7.0, and
sequence-parallel size 8.

\paragraph{Phantom.}
We use the released Phantom-14B subject-conditioned model for shots with references.
Each video reference is represented by its first native frame, and image references
are assigned to their corresponding shots. This adaptation supplies appearance
information through the model's image-reference interface. Reference-conditioned shots use 40 UniPC steps
with text guidance 7.5, image guidance 5.0, and flow shift 5.0. Text-only shot generation
uses 50 steps with guidance 5.0 and flow shift 5.0.

\paragraph{RefAlign.}
We use the released RefAlign-14B checkpoint with its DINOv3 reference encoder. Each
video reference contributes its first native frame, while image references are supplied
directly. References are routed to the corresponding shot through the subject-image
interface. Image anchors condition their owning shots through this interface.
Each shot is generated with 50 steps, text guidance 5.0, reference guidance 7.5, and
flow shift 5.0.

\paragraph{StoryMem.}
We use the released memory-conditioned model and its visual-memory mechanism. Each
video reference contributes its first, middle, and last native frames, with duplicate
positions removed for short clips. Image references are added directly to the initial
memory. These images form global visual memory available across shots. After each
shot, generated keyframes are added to the memory, retaining the initial references
and the most recent generated frames within a ten-image limit. Memory-conditioned
generation uses 40 UniPC steps, guidance 3.5, and flow shift 4.0. For text-only requests,
Wan2.2-T2V-A14B generates the first shot, and subsequent shots use generated memory.
Native 16-fps outputs are mapped to the requested 24-fps timeline by nearest-frame
selection before concatenation.

\paragraph{VACE.}
We evaluate Wan2.1-VACE-14B on the image- and video-conditioned requests using sparse
reference placement. Native reference frames are placed at selected positions in each
shot's input canvas without temporal interpolation. Image anchors occupy their assigned
positions. Occupied pixels are marked as known regions, and the remaining positions
are marked for generation. Sampling uses 50 UniPC steps, guidance 5.0, and flow shift
16.0. The internal canvas respects the implementation's 480p area and sequence-length
limits, and generated shots are spatially fitted to the requested output geometry before
concatenation. Reference Semantic, Subject, and Background are left unreported for this
completion-based adaptation, in which supplied reference regions are reconstruction
conditions.

\paragraph{Metric reporting and computational cost.}
Crosses in Table~\ref{tab:fused300-main} indicate unreported or inapplicable measurements.
Transition MAE is omitted for methods whose boundaries are fixed by shot concatenation.
Reference Semantic, Subject, and Background are unreported for VACE and the pretrained
HunyuanVideo1.5 baseline. All reported SDC scores use the same real-video calibration
bank. The real target videos of the benchmark reach SDC 0.332.

FLOPs are analytical denoising estimates for a $480\times832$, 241-frame video with
three shots and three 49-frame references. The estimates include each method's sampling
schedule, guidance branches, reference processing within the denoiser, and active
adapters. Text and visual
encoders, VAE processing, data transfer, and distributed communication are excluded
consistently. Sequence parallelism partitions the computation without reducing its
total FLOPs.

\section{SDC Details and Validation}
\label{app:sdc}

\paragraph{Sampling and representations.}
\label{sec:sdc-appendix}
Both generated and real videos use TransNetV2 shot boundaries. We remove three frames
at each boundary and sample two to four distinct, uniformly spaced interior frames per shot.
SDC requires at least two usable shots. Content uses DINOv2-small CLS embeddings, and
style uses the CSD ViT-L/14 style head and VGG16~\citep{simonyan2015vgg} channel means and standard deviations
from \texttt{relu1\_2}, \texttt{relu2\_2}, \texttt{relu3\_3}, and \texttt{relu4\_3}.
For VGG, each layer's concatenated moments are normalized before concatenation across layers.
All final frame embeddings are $L_2$-normalized.

\paragraph{Energy contrast and closed form.}
For frame features $x^r_{i,a}$, define $(D^r_{ij})_{ab}=1-\langle x^r_{i,a},x^r_{j,b}\rangle$
and the shot mean $\mu_i^r=n_i^{-1}\sum_a x^r_{i,a}$.
Distance averages include every entry, including the zero within-shot diagonals. Thus,
\begin{equation}
\begin{aligned}
    E^r_{ij}
      &=2\operatorname{mean}(D^r_{ij})
        -\operatorname{mean}(D^r_{ii})-\operatorname{mean}(D^r_{jj})\\
      &=2(1-\langle\mu^r_i,\mu^r_j\rangle)
        -(1-\|\mu^r_i\|^2)-(1-\|\mu^r_j\|^2)\\
      &=\|\mu^r_i-\mu^r_j\|^2.
\end{aligned}
\label{eq:sdc-v5-energy}
\end{equation}
Subtracting the within-shot distance terms gives a comparison of sample means,
equivalent to empirical squared MMD with a linear kernel
\citep{szekely2013energy,gretton2012mmd,sejdinovic2013equivalence}.
Floating-point round-off is handled by setting energy values in
$[-8\epsilon_{32},0)$ to zero, where $\epsilon_{32}$ denotes float32 machine precision.
Values below $-8\epsilon_{32}$ are rejected as invalid.

\paragraph{Within-shot calibration.}
To measure natural variation without introducing a shot transition, we partition each
shot's sampled frames into balanced, disjoint groups and compute their content and style
energies. Each unordered split is used once: two, three, and four frames yield one,
three, and three splits, respectively. The bank contains 2{,}144{,}201 real videos,
6{,}462{,}692 shots, and 19{,}388{,}074 splits, with all three representations available
on the same frames. All but six shots have four sampled frames.

For four sampled frames, define $d_a=x^r_{i,a}-\mu_i^r$, which satisfies $\sum_a d_a=0$.
The three partitions $\Pi_i$ pair frame 1 with $j\in\{2,3,4\}$, giving
\begin{equation}
\begin{aligned}
    \frac{1}{3}\sum_{\pi\in\Pi_i}E^r_{i,\pi}
      &=\frac{1}{3}\sum_{j=2}^{4}\|d_1+d_j\|^2\\
      &=\frac{1}{3}\sum_{a=1}^{4}\|d_a\|^2
       =\operatorname{tr}(\widehat\Sigma_i^r).
\end{aligned}
\label{eq:sdc-variance-proof}
\end{equation}
Here $\widehat\Sigma_i^r$ is the sample covariance with divisor three. The average split
energy measures within-shot feature dispersion. For $n$ frames split into groups of
sizes $k$ and $n-k$, the average is $n\operatorname{tr}(\widehat\Sigma)/(k(n-k))$,
with covariance divisor $n-1$. The same factor applies to content and style features.

\paragraph{Calibration groups and source exclusion.}
The real-video bank is partitioned by source video into three disjoint calibration
subsets, indexed by $g\in\{1,2,3\}$. Each subset provides a separately fitted
content--style slope and conditional residual distribution. Averaging their ranks
combines the three calibration estimates into a style-head rank
(Section~\ref{sec:sdc-method}).

Source exclusion uses a separate five-fold partition to prevent a query's source
video from contributing to its calibration. After shuffling unique source videos
with seed 20260908, zero-based index $i$ determines group $g=1+(i\bmod3)$ and fold
$i\bmod5$. The query source's fold is excluded from both slope fitting and residual
ranking in every group. Outputs from all methods and edited variants of the same
source use the same excluded fold. Sources absent from the bank receive a fixed
fold based on their identifier. The two style heads, three groups, and five
excluded-fold configurations require 30 calibration fits, all using real videos.

\paragraph{Robust content correction.}
For each group $g$ and style head $r$, we fit style energy $s_b$ against content energy $c_b$ using Huber regression~\citep{huber1973robust}.
A bank split has weight $w_b=1/(N_g S_v B_{v,q})$, where $N_g$ counts the videos
remaining in group $g$ after source-fold exclusion, $S_v$ counts usable shots in
video $v$, and $B_{v,q}$ counts splits in shot $q$. These weights assign equal mass
to each video, distributed equally among its shots and then their splits.
Suppressing group, head, and excluded-fold indices, we initialize $\kappa$ with the
weighted median of $s_b/c_b$ for $c_b>\epsilon_{32}$ and iterate
\begin{equation}
\begin{aligned}
    z_b&=s_b-\kappa c_b,\qquad m=\operatorname{wmed}_{w}(z_b),\\
    \sigma&=1.4826\,\operatorname{wmed}_{w}(|z_b-m|),\\
    a_b&=\min\!\left(1,\frac{1.345\sigma}{\max(|z_b|,\epsilon_{64})}\right),\\
    \kappa_{\mathrm{new}}&=\frac{\sum_b w_b a_b c_b s_b}{\sum_b w_b a_b c_b^2}.
\end{aligned}
\label{eq:sdc-huber-app}
\end{equation}
The weighted median $m$ centers residuals for the scale estimate $\sigma$.
The through-origin fit uses uncentred residuals $z_b$ in its influence weights.
Here $\epsilon_{64}$ is float64 machine precision. Iteration stops at
$|\kappa_{\mathrm{new}}-\kappa|\leq10^{-12}(1+|\kappa|)$, after 100 iterations,
or when $\sigma\leq\epsilon_{64}$.
Query residuals $e^r_{ij,g}=E^r_{ij}-\kappa_{g,r}E^c_{ij}$ retain their sign.
The residual correction subtracts the fitted linear content--style trend.
At fixed content energy, subtracting the same linear term preserves style ordering.
Queries outside the range of bin centers in content-percentile space use the nearest bin.

\paragraph{Responses by edit type.}
\label{app:sdc-population}
In the single-shot style-edit comparison in Table~\ref{tab:sdc-population}(a),
one shot receives a style-transfer edit and the other remains unedited,
with the corresponding original shot pair serving as the control.
The shared-versus-independent comparison in Table~\ref{tab:sdc-population}(b)
uses the same two reference-shot positions in both conditions, with both shots edited.
Comparisons retain matched shot pairs with valid scores for all compared metrics.
Scores are averaged over eligible pairs within each video and then equally across videos.
For each metric, the fifth percentile of the original-video scores in (a) or the
shared-edit scores in (b) defines the detection threshold.
The reported detection rate is the fraction of edited videos scoring below this threshold.

Table~\ref{tab:sdc-edit-categories} separates the responses by edit type.
SDC decreases more for style, filter, and lighting edits than for motion or spatial
changes.

\begin{table}[htbp]
\centering\small
\caption{\textbf{Sensitivity to reference edit types.} Relative decrease in mean similarity or consistency from original shot pairs to pairs with one edited shot. Larger decreases indicate stronger responses within each metric's scale. A source video can appear in multiple edit categories.}
\label{tab:sdc-edit-categories}
\setlength{\tabcolsep}{5pt}
\begin{tabular*}{\linewidth}{@{\extracolsep{\fill}}lrrrrr@{}}
\toprule
Edit type & Videos & SDC & CLIP & ViCLIP & DINO \\
\midrule
Camera / lens filter & 21,699 & 96.2\% & 13.4\% & 33.9\% & 22.0\% \\
Style transfer & 108,909 & 86.5\% & 14.9\% & 30.1\% & 22.8\% \\
Lighting / weather & 29,516 & 75.0\% & 10.4\% & 22.5\% & 17.1\% \\
Background retexture & 27,253 & 64.2\% & 10.2\% & 24.4\% & 21.0\% \\
Object material / color & 23,405 & 51.0\% & 7.8\% & 16.6\% & 10.1\% \\
Subject appearance & 29,873 & 32.4\% & 9.5\% & 18.0\% & 13.5\% \\
Motion change & 22,559 & 19.2\% & 4.6\% & 7.1\% & 3.8\% \\
Spatial relation change & 20,424 & 18.6\% & 4.0\% & 6.3\% & 3.2\% \\
\bottomrule
\end{tabular*}
\end{table}

\paragraph{Comparison with style-feature similarity.}
Table~\ref{tab:sdc-style-similarity} compares SDC with CSD and VGG cosine similarity
using the same frame features and video-level averaging.
Each cosine score averages all cross-shot frame pairs of normalized style features.
The edit comparisons follow the control thresholds defined above.
For the real-video comparison, each metric's fifth percentile over all 2,144,201
real videos defines a low-score threshold. The last column reports the fraction of
the 458,920 videos with DINO similarity below 0.25 falling below that threshold.
CSD achieves the highest detection rates in both edit comparisons.
SDC retains sensitivity to style conflicts while reducing score dependence on content changes,
with a low-score rate of 4.4\% on the low-DINO subset compared with 18.8\% for CSD
and 13.7\% for VGG.

\begin{table}[htbp]
\centering\small
\caption{\textbf{Style-feature similarity and SDC.} Edit detection uses a 5\% control
false-positive rate. Low-score rates use each metric's fifth percentile over all
real videos as the threshold.}
\label{tab:sdc-style-similarity}
\setlength{\tabcolsep}{6pt}
\begin{tabular*}{\linewidth}{@{\extracolsep{\fill}}lccc@{}}
\toprule
Metric & \shortstack{Single-shot style edits\\Detection $\uparrow$}
& \shortstack{Independent vs. shared edits\\Detection $\uparrow$}
& \shortstack{Real videos, DINO $<0.25$\\Low-score rate} \\
\midrule
CSD cosine & 78.6\% & 58.5\% & 18.8\% \\
VGG cosine & 34.9\% & 39.5\% & 13.7\% \\
SDC & 78.0\% & 50.9\% & 4.4\% \\
\bottomrule
\end{tabular*}
\end{table}

\paragraph{Effect of linear correction.}
\label{app:sdc-kappa-ablation}
We set $\kappa=0$ for both bank and query energies in both style heads, keeping the
features, calibration videos, source-fold exclusion, bin boundaries, sample weights,
interpolation, and fusion fixed. Within each bin, bank samples are reordered by
uncorrected style energy. The comparison reuses the 108,909 single-shot style-edit
videos, 72,570 shared/independent-edit videos, and 2,144,201 real videos above.
Thresholds follow the preceding comparison and are recalculated for each variant.

\begin{table}[htbp]
\centering\small
\caption{\textbf{Effect of linear correction with fixed conditional ranking.}
Edit detection uses a 5\% control false-positive rate. The last column gives the
low-score fraction among 458,920 real videos with DINO $<0.25$, using each
variant's fifth percentile over all real videos as its threshold.}
\label{tab:sdc-kappa-ablation}
\setlength{\tabcolsep}{6pt}
\begin{tabular*}{\linewidth}{@{\extracolsep{\fill}}lccc@{}}
\toprule
Setting & \shortstack{Single-shot style edits\\Detection $\uparrow$}
& \shortstack{Independent vs. shared edits\\Detection $\uparrow$}
& \shortstack{Real videos, DINO $<0.25$\\Low-score rate} \\
\midrule
$\kappa=0$ & 76.79\% & 50.29\% & 6.01\% \\
Full SDC & 77.98\% & 50.85\% & 4.38\% \\
\bottomrule
\end{tabular*}
\end{table}

Linear correction increases the two edit-detection rates by 1.19 and 0.56 percentage
points (Table~\ref{tab:sdc-kappa-ablation}). Their 95\% intervals are [0.95, 1.42] and
[0.23, 0.82] percentage points, respectively. The low-DINO real-video low-score rate
decreases by 1.63 points, with a 95\% interval of [1.60, 1.67]. Intervals use 2,000
paired source-video bootstrap draws, re-estimating each variant's threshold in every
draw while holding the calibration bank fixed.

For real-video shot pairs, the mean absolute score change is 0.0060 when ranks
interpolate between bin centers and 0.0728 when a query uses an endpoint bin in at
least one calibration group or style head. Across all real videos, Spearman
correlation with DINO is $-0.0879$ with correction and $+0.0184$ at $\kappa=0$.
Linear correction gives modest detection gains, while $\kappa=0$ has a global
content correlation closer to zero. The largest score changes occur for queries
using endpoint bins.

\subsection{Calibration Stability}
\label{app:kappa}

Table~\ref{tab:kappa-scaling} evaluates six sample sizes from 100 to all 2{,}144{,}201
usable real videos. We refit the CSD/DINOv2 slope on each subset using
Equation~\ref{eq:sdc-huber-app} and the bank's video--shot--split weights.
Each size has five disjoint subsets up to 100{,}000 videos, two at one million,
and one at the full size. Subsets at different sizes can overlap.
SDC is averaged over these same query subsets using the fixed full real-video bank
with source-fold exclusion.

Mean $\kappa$ ranges from 0.39100 to 0.40168, reaching 0.39874 on the full pool.
Its standard deviation across repeated fits decreases from 0.01533 at 100 videos
to 0.00014 at one million. Mean SDC ranges from 0.33636 to 0.34414 and is 0.33731
on the full pool. The two columns measure slope-estimation stability and the sampling
stability of mean SDC under a fixed bank, respectively. With IP and Stock added, the pooled 86.9M-video bank gives $\kappa=0.41092$ and a mean real-video SDC of 0.32644 (last row).

\begin{table}[htbp]
\centering\small
\caption{\textbf{Real-video $\kappa$ and SDC across six sample sizes.}
Repeated subsets are disjoint within each size. We refit $\kappa$ on each subset
and average SDC over the same videos using the fixed full real-video bank.
Standard deviations and ranges summarize repetitions, and the 2,144,201 row uses all videos. The last row is the pooled bank with IP and Stock, where $\kappa$ is fitted and SDC is scored on that bank.}
\label{tab:kappa-scaling}
\begingroup
\setlength{\tabcolsep}{4pt}
\begin{tabular}{@{}rrrrrrrr@{}}
\toprule
 & & \multicolumn{4}{c}{$\kappa$} & \multicolumn{2}{c}{SDC} \\
\cmidrule(lr){3-6}\cmidrule(l){7-8}
Videos & Repeats & Mean & sd & Minimum & Maximum & Mean & sd \\
\midrule
100 & 5 & 0.39100 & 0.01533 & 0.36996 & 0.41107 & 0.33636 & 0.02372 \\
1,000 & 5 & 0.40168 & 0.00284 & 0.39684 & 0.40415 & 0.34414 & 0.00484 \\
10,000 & 5 & 0.39920 & 0.00065 & 0.39813 & 0.39984 & 0.33795 & 0.00227 \\
100,000 & 5 & 0.39890 & 0.00066 & 0.39793 & 0.39982 & 0.33711 & 0.00071 \\
1,000,000 & 2 & 0.39872 & 0.00014 & 0.39858 & 0.39886 & 0.33733 & 0.00002 \\
2,144,201 & 1 & 0.39874 & --- & 0.39874 & 0.39874 & 0.33731 & --- \\
\midrule
86,908,057 & 1 & 0.41092 & --- & 0.41092 & 0.41092 & 0.32644 & --- \\
\bottomrule
\end{tabular}
\endgroup
\end{table}

With the source population, encoders, and sampling protocol fixed, larger samples
estimate the same content--style relationship more precisely. The real-video bank is
a stable reference for comparison. All generators use
this shared bank for both slope correction and conditional ranking. To test whether this conclusion depends on the choice of bank, we rebuild the calibration from five banks that differ in source and size (Table~\ref{tab:sdc-bank-scaling}). Three banks contain the 0.22M DurMix source videos in their original, shared-edit, and independent-edit versions. The paper bank contains the 2.14M web videos collected for this work. The pooled bank adds two separately collected real-video collections to the paper bank, giving 86.9M videos and about 1.0B within-shot splits. IP contains 17.2M clips from films, TV series, anime, and game cinematics. Stock contains 67.6M everyday web videos such as tutorials, cooking, travel, and aerial footage. Across a 400-fold increase in bank size, the content--style slope remains within 0.388--0.411 and the ordering of all six methods is identical under every bank (Kendall $\tau=1.0$).
\begin{table}[h]
\centering\scriptsize
\caption{\textbf{SDC under five calibration banks.} Mean SDC on the benchmark, with each bank used for both slope fitting and conditional ranking under source-fold exclusion. The pooled bank combines the paper bank with IP and Stock (86.9M videos, about 1.0B within-shot splits). Kendall $\tau$ compares each method ranking with the paper bank.}
\label{tab:sdc-bank-scaling}
\setlength{\tabcolsep}{3pt}
\begin{tabular}{lrccccccc}
\toprule
Calibration bank & Videos & $\hat\kappa_{\rm CSD}$ & Refrain & StoryMem & Phantom & RefAlign & VACE & HY1.5 \\
\midrule
DurMix original & 0.22M & 0.388 & \textbf{0.291} & 0.160 & 0.114 & 0.110 & 0.094 & 0.080 \\
DurMix shared edits & 0.22M & 0.393 & \textbf{0.306} & 0.167 & 0.127 & 0.123 & 0.105 & 0.087 \\
DurMix indep.\ edits & 0.22M & 0.393 & \textbf{0.305} & 0.167 & 0.126 & 0.122 & 0.104 & 0.086 \\
Paper bank & 2.14M & 0.399 & \textbf{0.315} & 0.175 & 0.130 & 0.125 & 0.108 & 0.090 \\
Pooled real & 86.9M & 0.411 & \textbf{0.277} & 0.158 & 0.102 & 0.099 & 0.084 & 0.076 \\
\bottomrule
\end{tabular}
\par\smallskip
Kendall $\tau=1.0$ for every bank.
\end{table}

\Needspace{10\baselineskip}
\section{Human Study}
\label{app:human}

\begin{figure}[htbp]
\centering
\includegraphics[width=0.78\linewidth]{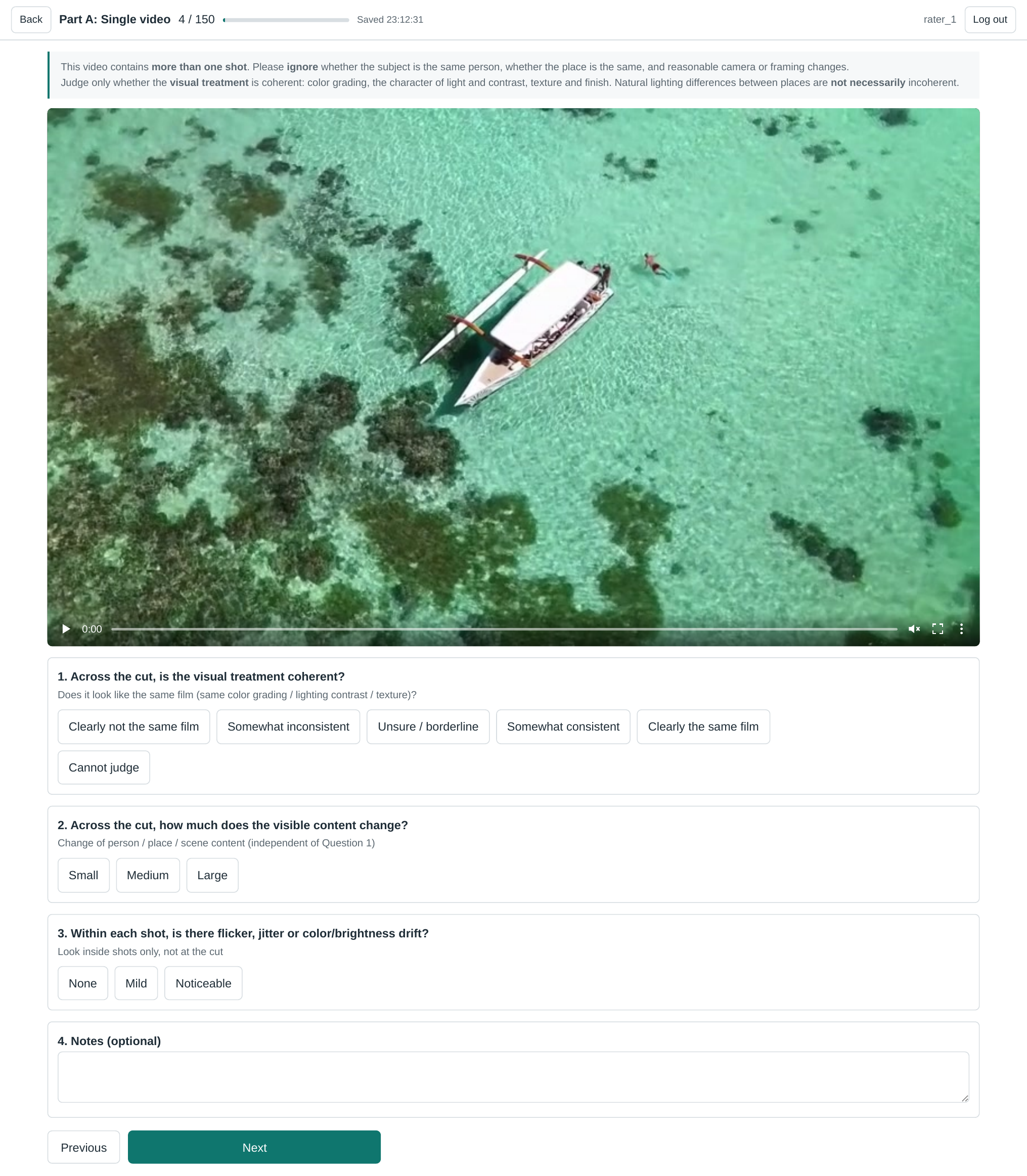}\par\small (a) Full page\par\medskip
\includegraphics[width=\linewidth,trim=40 161 40 620,clip]{figure/user_study_A.pdf}\par\small (b) Rating questions, enlarged
\caption{\textbf{Rating interface.} Raters judge whether the visual treatment is coherent across the cut (colour grading, lighting, contrast, texture) on five levels, and ignore changes of subject, place, or framing. They also rate the amount of content change and within-shot flicker, and may mark ``cannot judge''.}
\label{fig:us-interface-a}
\end{figure}

\paragraph{Design.} The protocol, sample, and analysis plan were frozen before any rating was collected. Six professional raters, unaffiliated with this project, rated 150 single clips in an order randomized per rater, with method identities and SDC scores hidden. The set contains 60 real two-shot splices joining shots from different source videos, drawn from our sources and from the IP and Stock collections, which exclude every training and test video by file hash. It also contains 30 natural real multi-shot clips, 40 generated clips from four video generators, 12 further clips (six Refrain-generated clips, five splices of two independently edited shots from the same source video, and one real clip), and 8 test--retest repeats. Splices combine the Stock and IP collections in all pairings, and pairs are either matched by shot-text similarity or random. Content relatedness and style compatibility therefore vary independently. Text-matched shots can still clash in style, and shots from different sources can still be compatible. Raters judged colour grading, lighting, contrast, and texture across the cut on a five-level scale, from 1 (clearly inconsistent style) through 3 (borderline) to 5 (clearly consistent style), and were instructed to ignore changes of subject, place, or framing (Figure~\ref{fig:us-interface-a}).

\paragraph{Consistency labels and scoring.} Ratings of 1--2 indicate inconsistency, 4--5 consistency, and 3 is undecided. A clip receives a class label when at least four of six raters agree. All clips are scored with the same real-video bank and CSD/VGG fusion as the main evaluation, with source-fold exclusion, and spliced clips are scored on the videos shown to raters at their recorded splice boundaries. AUC measures the probability that a consensus-consistent clip receives a higher SDC than a consensus-inconsistent clip, with half credit for ties. Excluding the 8 repeats and 2 clips in which TransNetV2 detects no cut, 140 clips are scored: 73 consistent, 48 inconsistent, and 19 undecided.

\begin{figure}[htbp]
\centering
\includegraphics[width=0.78\linewidth]{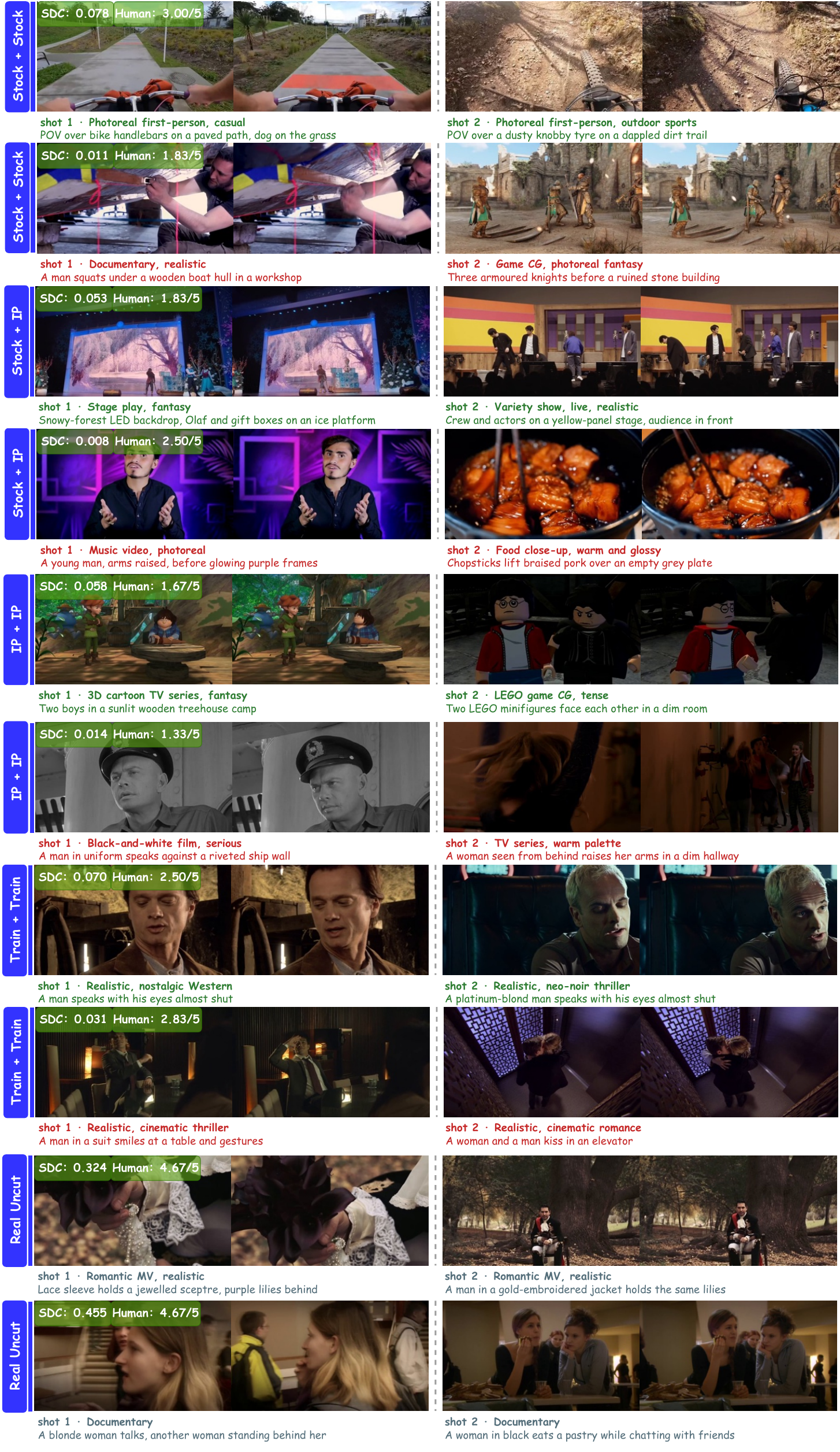}
\caption{\textbf{SDC and human ratings on real footage.} Each row is one rated clip: two-shot splices from different sources (Stock + Stock, Stock + IP, IP + IP, Train + Train) and natural multi-shot clips (Real Uncut), with the shot descriptions, SDC, and the mean score of six raters (1--5).}
\label{fig:sdc-splices}
\end{figure}

\paragraph{Agreement and SDC.} Krippendorff's ordinal $\alpha$ is 0.68. Fleiss' $\kappa$ is 0.35 on the five-level scale and 0.50 after grouping ratings into three classes. Over all 140 clips, SDC agrees with the mean rating at $\rho=0.68$ (95\% interval [0.59, 0.76]) and reaches AUC 0.90 ([0.84, 0.95]), with intervals from 20{,}000 clip-level bootstrap draws. Mean SDC is 0.327 for consistent clips and 0.057 for inconsistent clips. On the 90 real clips (40 consistent, 36 inconsistent), SDC agrees with the raters at $\rho=0.77$ and AUC 0.97, on independently collected footage as well as on training-source footage. On the 54 real clips from the two external collections alone, SDC agrees at $\rho=0.79$ and AUC 0.98 (Table~\ref{tab:sdc-human-strata}). Some splices of different sources are also judged consistent. Figure~\ref{fig:sdc-splices} shows rated examples, including style-conflicting splices with low SDC and low ratings and natural multi-shot clips rated as consistent.

\begin{table}[h]
\centering\footnotesize
\caption{\textbf{SDC and human ratings by subset}. $n$: clips (consistent/inconsistent, rest undecided). $\rho$: Spearman. Ext.: the two external collections. Train: our source videos.}
\label{tab:sdc-human-strata}
\setlength{\tabcolsep}{3.5pt}
\begin{tabular}{lcccccc}
\toprule
 & All & Real & Real, ext. & Ext.\ splices & Real, train & Generated \\
\midrule
$n$ & 140 (73/48) & 90 (40/36) & 54 (25/22) & 36 (8/21) & 36 (15/14) & 38 (25/8) \\
$\rho$ & 0.68 & 0.77 & 0.79 & 0.59 & 0.75 & 0.30 \\
AUC & 0.90 & 0.97 & 0.98 & 0.95 & 0.97 & 0.57 \\
\bottomrule
\end{tabular}
\end{table}

\section{Cascade Baseline}
\label{app:cascade}
A cascade can first restyle every reference toward the shared style prompt and then generate from the restyled references. Every video reference is restyled with FLUX.1 Kontext on the first frame and depth-conditioned Wan-2.1 propagation, image anchors are edited directly, and the edited references pass the same filters as \durmix{}. They are then given to Refrain with unchanged prompts. Table~\ref{tab:cascade} reports the requests with video references. The cascade leaves SDC nearly unchanged (0.324 vs.\ 0.323) while lowering reference subject and background fidelity, TA, and transition timing. Measured against the restyled references that it received, the cascade's subject fidelity rises to 0.627 and remains below that of Refrain on the original references (0.645). Restyling the 639 image and video references took about 73 GPU-hours including regeneration, about 7 GPU-minutes per reference, most of it in Wan-2.1 propagation (median 3.8 GPU-minutes per video reference). Of these references, 100 failed the filters and were regenerated, 67 of them twice. The filtering time is not included. Refrain reaches the same consistency in one pass without this preprocessing and keeps more of the reference content.
\begin{table}[h]
\centering\scriptsize
\caption{\textbf{Cascade baseline} (requests with video references). References are restyled with the \durmix{} editors before generation. Fidelity is measured against the original references and, for the cascade rows, also against the restyled references the generator received. Real videos of the same requests: SDC 0.339.}
\label{tab:cascade}
\setlength{\tabcolsep}{4pt}
\begin{tabular}{lccccccc}
\toprule
 & & \multicolumn{2}{c}{vs.\ original refs} & \multicolumn{2}{c}{vs.\ restyled refs} & & \\
\cmidrule(lr){3-4}\cmidrule(lr){5-6}
Method & SDC$\uparrow$ & Ref-Subj$\uparrow$ & Ref-Bg$\uparrow$ & Ref-Subj$\uparrow$ & Ref-Bg$\uparrow$ & TA$\uparrow$ & MAE (s)$\downarrow$ \\
\midrule
Refrain & 0.323 & \textbf{0.645} & \textbf{0.653} & -- & -- & \textbf{$-$0.204} & \textbf{0.197} \\
Cascade + Refrain & 0.324 & 0.569 & 0.574 & 0.627 & 0.638 & $-$0.296 & 0.227 \\
\bottomrule
\end{tabular}
\end{table}

\section{Generalization of the Training Edits}
\label{app:generalization}
\paragraph{Hybrid training.} Hybrid training draws shared or independent edits with probability 0.5 per video, and all other training settings are identical (Table~\ref{tab:hybrid}). It matches shared-edit training on both test sets and trails independent-edit training on independent edits. Independent-edit training exceeds shared-edit training by 0.014 SDC on independent edits and is close to it on shared edits (0.299 vs.\ 0.304). The gain comes from the independent supervision, in which references of one video carry different edits and the model learns to reconcile them. In the mixture, half of the videos carry one shared edit and provide no such conflict, and the results fall to those of shared-edit training.
\begin{table}[h]
\centering\scriptsize
\caption{\textbf{Hybrid edit training} (SDC).}
\label{tab:hybrid}
\begin{tabular}{lcc}
\toprule
Training edits & Independent-edit test & Shared-edit test \\
\midrule
Independent & \textbf{0.315} & 0.299 \\
Shared & 0.301 & 0.304 \\
Hybrid 1:1 & 0.300 & 0.303 \\
\bottomrule
\end{tabular}
\end{table}

\begin{figure}[htbp]
\centering
\includegraphics[width=\linewidth]{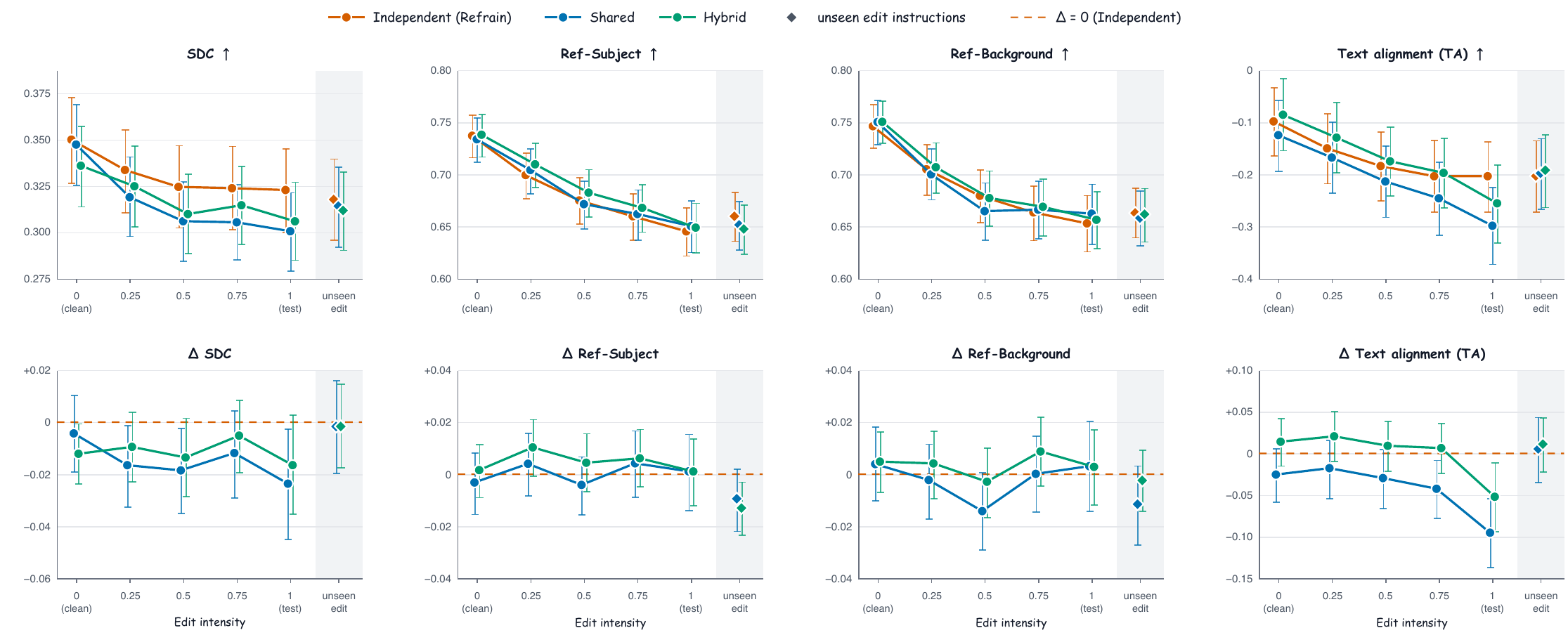}
\caption{\textbf{Performance curves vs.\ edit intensity.} Top: SDC, Ref-Subject, Ref-Background, and TA of the independent-edit (Refrain), shared-edit, and hybrid training, from clean references ($\alpha{=}0$) to the benchmark edits ($\alpha{=}1$), plus unseen edit instructions (shaded). Bottom: Shared minus Independent and Hybrid minus Independent, averaged over the requests scored in both settings. The top row and Table~\ref{tab:generalization} average each setting over all of its scored requests, and the two averages can differ slightly.}
\label{fig:performance-curve}
\end{figure}

\begin{figure}[htbp]
\centering
\noindent\makebox[0.04\linewidth]{}\makebox[0.416\linewidth]{\small References}\makebox[0.544\linewidth]{\small Refrain outputs}\par\smallskip
\includegraphics[width=\linewidth]{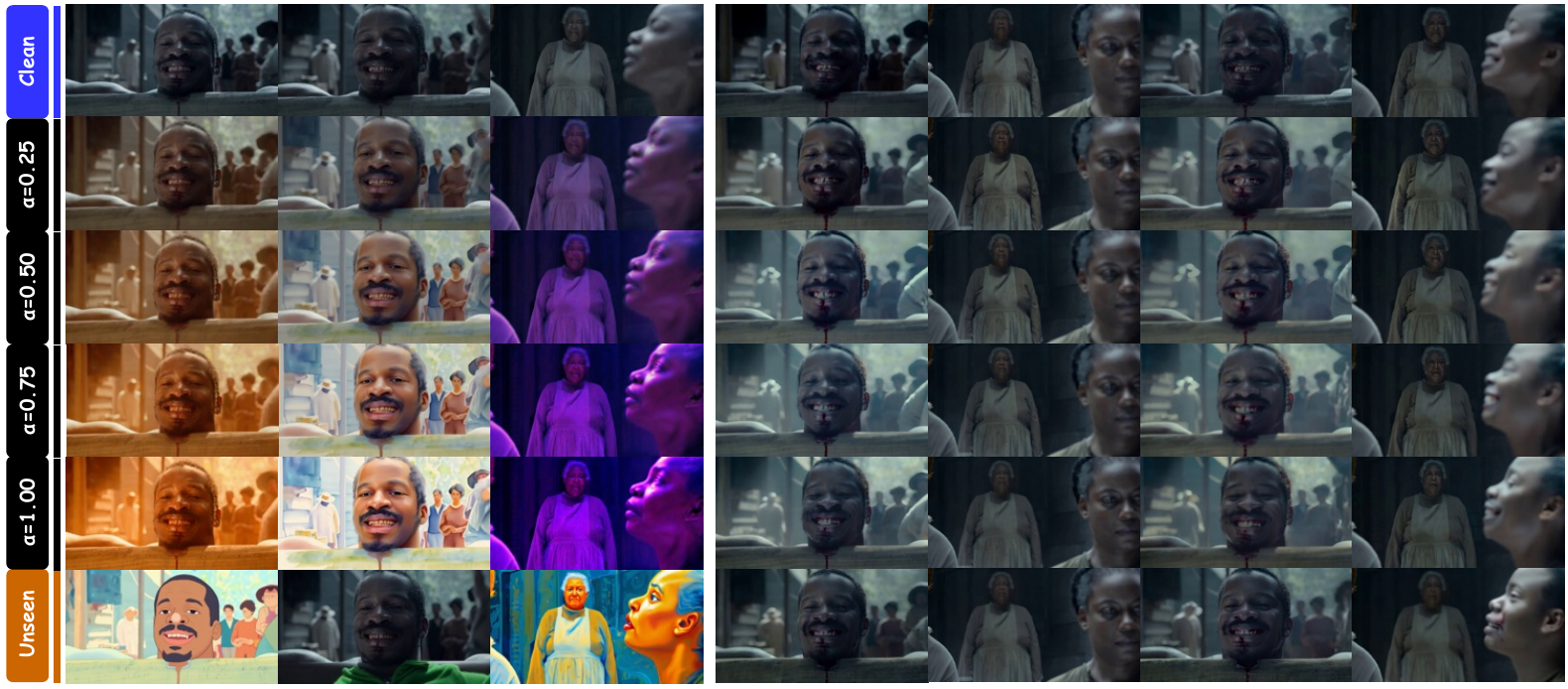}
\caption{\textbf{One request under six reference conditions.} Left: the references given to the model (clean, blends $\alpha=0.25$--$1$, unseen instructions). Right: Refrain's output frames for each condition.}
\label{fig:performance-examples}
\end{figure}

\paragraph{Performance curves.} We vary the test references and evaluate all three training settings on the requests with video references (Figure~\ref{fig:performance-curve}, Table~\ref{tab:generalization}). \emph{Clean}: the original, unedited references. \emph{Blend}: per-frame RGB interpolation $(1-\alpha)R_i+\alpha\widetilde R_i$ between original and edited references, re-encoded, where $\alpha$ controls edit intensity. \emph{Unseen instructions}: 20 newly written instructions absent from all training data (e.g., gouache painting, colored-pencil illustration, 1980s camcorder look, misty blue-hour dawn, faded instant film, emerald-green recolouring), passed through the same filters. Independent-edit training gives the highest SDC at every intensity and under unseen instructions. The three settings have similar SDC on clean references, and the gap opens once the references conflict. On the blended references, hybrid training lies between the other two. Figure~\ref{fig:performance-examples} shows one request under all six conditions. The references change from clean to strongly stylized, and Refrain's outputs keep one consistent look.

\begin{table}[h]
\centering\scriptsize
\caption{\textbf{Performance under varying test references} (requests with video references). $\alpha$: blend weight between original and edited references ($\alpha{=}0$ clean, $\alpha{=}1$ benchmark edits). Fidelity is measured against the references each run received.}
\label{tab:generalization}
\setlength{\tabcolsep}{3pt}
\begin{tabular}{lccc|c}
\toprule
 & \multicolumn{3}{c|}{SDC$\uparrow$} & Ref-Subj$\uparrow$ \\
Test references & Independent & Shared & Hybrid & Independent \\
\midrule
Clean ($\alpha{=}0$) & \textbf{0.350} & 0.347 & 0.336 & 0.737 \\
$\alpha{=}0.25$ & \textbf{0.334} & 0.319 & 0.325 & 0.699 \\
$\alpha{=}0.50$ & \textbf{0.324} & 0.306 & 0.310 & 0.675 \\
$\alpha{=}0.75$ & \textbf{0.324} & 0.305 & 0.315 & 0.660 \\
Benchmark ($\alpha{=}1$) & \textbf{0.323} & 0.301 & 0.306 & 0.645 \\
Unseen instructions & \textbf{0.318} & 0.314 & 0.312 & 0.660 \\
\bottomrule
\end{tabular}
\end{table}

\section{Channel vs.\ Sequence Injection}
\label{app:rebuttal-exps}
Both variants are trained on the same data with identical settings, and only the injection differs. In the sequence variant, reference latents are appended as extra tokens with shot-aware temporal coordinates at their target intervals. Both are evaluated under the protocol of Table~\ref{tab:fused300-main} (Table~\ref{tab:seq-concat}). The two variants are close on every metric. SDC is 0.315 for both, and the largest gaps are in TA ($-$0.250 vs.\ $-$0.221) and inter-shot subject consistency (0.476 vs.\ 0.494).

\begin{table}[h]
\centering\scriptsize
\caption{\textbf{Channel vs.\ sequence injection} (metrics of Table~\ref{tab:fused300-main}).}
\label{tab:seq-concat}
\setlength{\tabcolsep}{4pt}
\begin{tabular}{lccccccccc}
\toprule
 & & & & & \multicolumn{3}{c}{Inter-Shot$\uparrow$} & \multicolumn{2}{c}{Reference$\uparrow$} \\
\cmidrule(lr){6-8}\cmidrule(lr){9-10}
Injection & MAE\,s$\downarrow$ & CLIP-T$\uparrow$ & TA$\uparrow$ & SDC$\uparrow$ & Semantic & Subject & Scene & Subject & Background \\
\midrule
Channel (Refrain) & 0.199 & 0.222 & $-$0.221 & 0.315 & 0.710 & 0.494 & 0.491 & 0.671 & 0.674 \\
Sequence & 0.209 & 0.222 & $-$0.250 & 0.315 & 0.709 & 0.476 & 0.493 & 0.666 & 0.670 \\
\bottomrule
\end{tabular}
\end{table}

\paragraph{Cost.} Channel concatenation keeps the token count of the target for any number and length of references. A 241-frame target has 61 latent frames, and three 49-frame references add $3\times13=39$, and the sequence variant processes 1.64$\times$ as many tokens. Self-attention cost grows with the square of this count (about 2.7$\times$), and the analytical denoising cost under the configuration of Table~\ref{tab:fused300-main} rises from 324 to 811 PFLOPs per request (2.5$\times$). With identical settings, a denoising step takes 24.7\,s for the sequence variant against 13.8\,s for the channel variant (1.8$\times$), and the median training update takes 28.4\,s against 14.2\,s (2.0$\times$). This overhead grows with every added reference.

\section{Limitations}
Refrain is primarily evaluated on short multi-shot sequences, and maintaining recurring identities and a coherent style across substantially longer narratives requires further evaluation.
The current task uses a shared style prompt to specify the desired visual appearance across shots, while deliberate shot-specific style changes require a more detailed description of the intended output.
Explicit control over which reference attributes to preserve or revise remains a direction for future work.
DurMix-300K pairs edited references with their original videos, and its coverage is bounded by the source footage and editing transformations used in construction.
SDC measures cross-shot style consistency relative to a fixed feature space and calibration bank, and its scores should be interpreted alongside reference fidelity, motion quality, and human judgments.

\section{Future Work}
A central direction is consistency-guided post-training, using SDC together with human preferences to improve cross-shot style consistency while preserving reference content and motion.
Comparing candidate videos generated from the same prompts and references could provide preference supervision, with reference fidelity and motion quality included in candidate selection.
We plan to examine whether this supervision improves consistency across all four conditioning tasks and how well these gains agree with human judgments.
We also aim to extend Refrain to longer narratives and references collected from independent videos, testing recurring subjects and locations across longer shot sequences and a broader range of visual domains.
Finally, we plan to explore localized revision, allowing users to replace a reference or revise an individual shot while preserving the timing, content, and style of the remaining sequence.

\section{Additional Qualitative Results}
\label{app:qualitative}

\subsection{Reference Duration and Motion Pacing}
\label{app:duration-examples}

Variable-duration reference control is a core capability of Refrain. The requested shot
can be longer or shorter than its reference, while the selected reference start and target
interval remain unchanged. Uniformly stretching or compressing the reference changes
motion speed, while truncating it can discard important actions. The examples below examine both
directions, showing how reference actions are adapted while preserving natural pacing and
a shared style across shots.

\paragraph{Results by reference duration.} Table~\ref{tab:duration-groups} groups the requests with video references by the duration of their video references relative to the assigned intervals. Refrain obtains the highest SDC, reference fidelity, and text alignment in all four groups, and the lowest transition error among the methods that generate the full sequence. Its SDC stays between 0.307 and 0.334 and its transition MAE between 0.166 and 0.208\,s across groups. For every method, requests that mix shorter and longer references give the lowest reference-subject similarity and text alignment.
\begin{table}[h]
\centering\scriptsize
\caption{\textbf{Results by reference duration} (requests with video references). Requests are grouped by the duration of their video references relative to the assigned intervals: all shorter, all equal, all longer, or a mix of shorter and longer references. $n$: requests per group. Bold: best per group. Transition MAE applies only to methods that generate all shots jointly.}
\label{tab:duration-groups}
\setlength{\tabcolsep}{6pt}
\begin{tabular}{lcccc}
\toprule
Method & Shorter ($n{=}55$) & Equal ($n{=}80$) & Longer ($n{=}26$) & Mixed ($n{=}79$) \\
\midrule
\multicolumn{5}{l}{\textit{SDC$\uparrow$}} \\
Phantom & 0.130 & 0.123 & 0.113 & 0.111 \\
RefAlign & 0.110 & 0.086 & 0.103 & 0.088 \\
StoryMem & 0.200 & 0.150 & 0.163 & 0.181 \\
VACE & 0.116 & 0.095 & 0.097 & 0.092 \\
\textbf{Refrain} & \textbf{0.334} & \textbf{0.329} & \textbf{0.329} & \textbf{0.307} \\
\midrule
\multicolumn{5}{l}{\textit{Ref-Subject$\uparrow$}} \\
Phantom & 0.496 & 0.431 & 0.429 & 0.413 \\
RefAlign & 0.485 & 0.469 & 0.470 & 0.453 \\
StoryMem & 0.419 & 0.410 & 0.426 & 0.353 \\
\textbf{Refrain} & \textbf{0.678} & \textbf{0.649} & \textbf{0.685} & \textbf{0.608} \\
\midrule
\multicolumn{5}{l}{\textit{Ref-Background$\uparrow$}} \\
Phantom & 0.377 & 0.363 & 0.348 & 0.379 \\
RefAlign & 0.474 & 0.405 & 0.477 & 0.469 \\
StoryMem & 0.359 & 0.398 & 0.422 & 0.372 \\
\textbf{Refrain} & \textbf{0.668} & \textbf{0.651} & \textbf{0.662} & \textbf{0.642} \\
\midrule
\multicolumn{5}{l}{\textit{Transition MAE (s)$\downarrow$}} \\
\textbf{Refrain} & \textbf{0.166} & \textbf{0.208} & \textbf{0.206} & \textbf{0.204} \\
\midrule
\multicolumn{5}{l}{\textit{TA$\uparrow$}} \\
Phantom & $-$0.649 & $-$0.698 & $-$0.632 & $-$0.791 \\
RefAlign & $-$0.514 & $-$0.525 & $-$0.436 & $-$0.636 \\
StoryMem & $-$0.727 & $-$0.705 & $-$0.733 & $-$0.920 \\
VACE & $-$0.292 & $-$0.257 & $-$0.254 & $-$0.404 \\
\textbf{Refrain} & \textbf{$-$0.239} & \textbf{$-$0.128} & \textbf{$-$0.081} & \textbf{$-$0.296} \\
\bottomrule
\end{tabular}
\end{table}

\paragraph{A shorter reference.}
Figure~\ref{fig:duration-short} (\texttt{case12\_v2v} in the supplementary videos)
uses a 3.7-second reference for a 6.5-second target shot. Refrain preserves the pace
of the camera gesture while bringing the referenced shot into the surrounding
shots' visual style.

\begin{figure}[htbp]
    \centering
    \includegraphics[width=\linewidth]{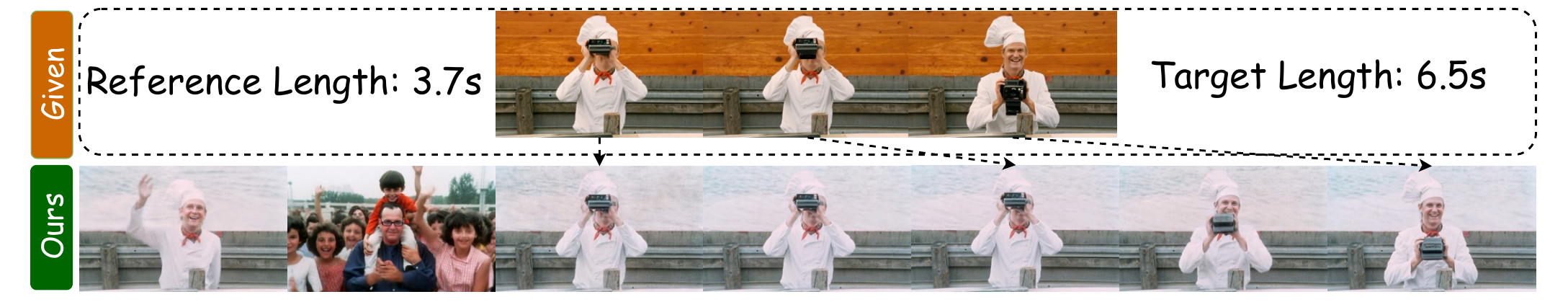}
    \caption{\textbf{Short reference, longer target: 3.7s $\to$ 6.5s.}
    Refrain preserves plausible action pacing and a shared style across shots. Dashed arrows link reference and generated action states.}
    \label{fig:duration-short}
\end{figure}

\paragraph{A longer reference.}
Figure~\ref{fig:duration-long} (\texttt{case13\_v2v} in the supplementary videos)
uses a 3.9-second reference for a 2.8-second target shot. Refrain condenses the action
sequence while retaining the woman's reaction, hand gesture, and subsequent head
turn at a natural pace.

\begin{figure}[htbp]
    \centering
    \includegraphics[width=\linewidth]{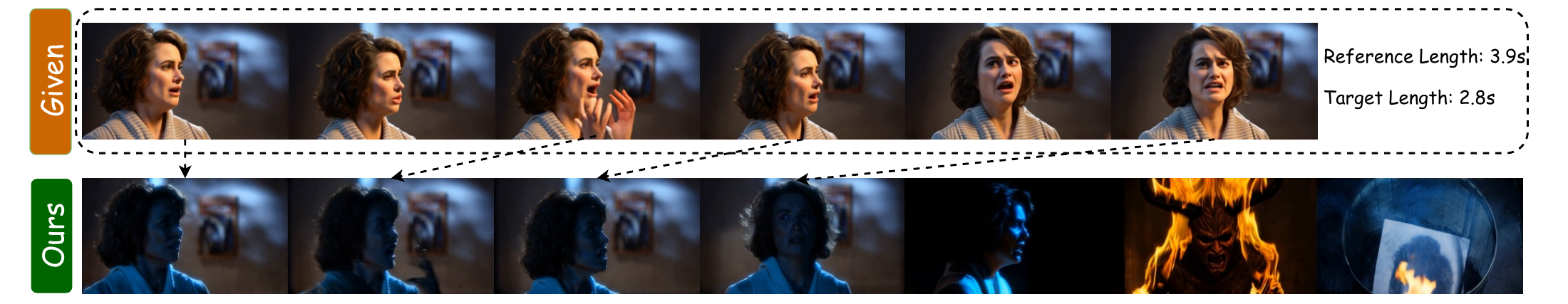}
    \caption{\textbf{Long reference, shorter target: 3.9s $\to$ 2.8s.}
    Refrain condenses the reference action while retaining its key gestures and
    natural pacing.}
    \label{fig:duration-long}
\end{figure}

\clearpage
\subsection{Further Generation Examples}

\begin{figure}[htbp]
    \centering
    \includegraphics[width=0.72\linewidth]{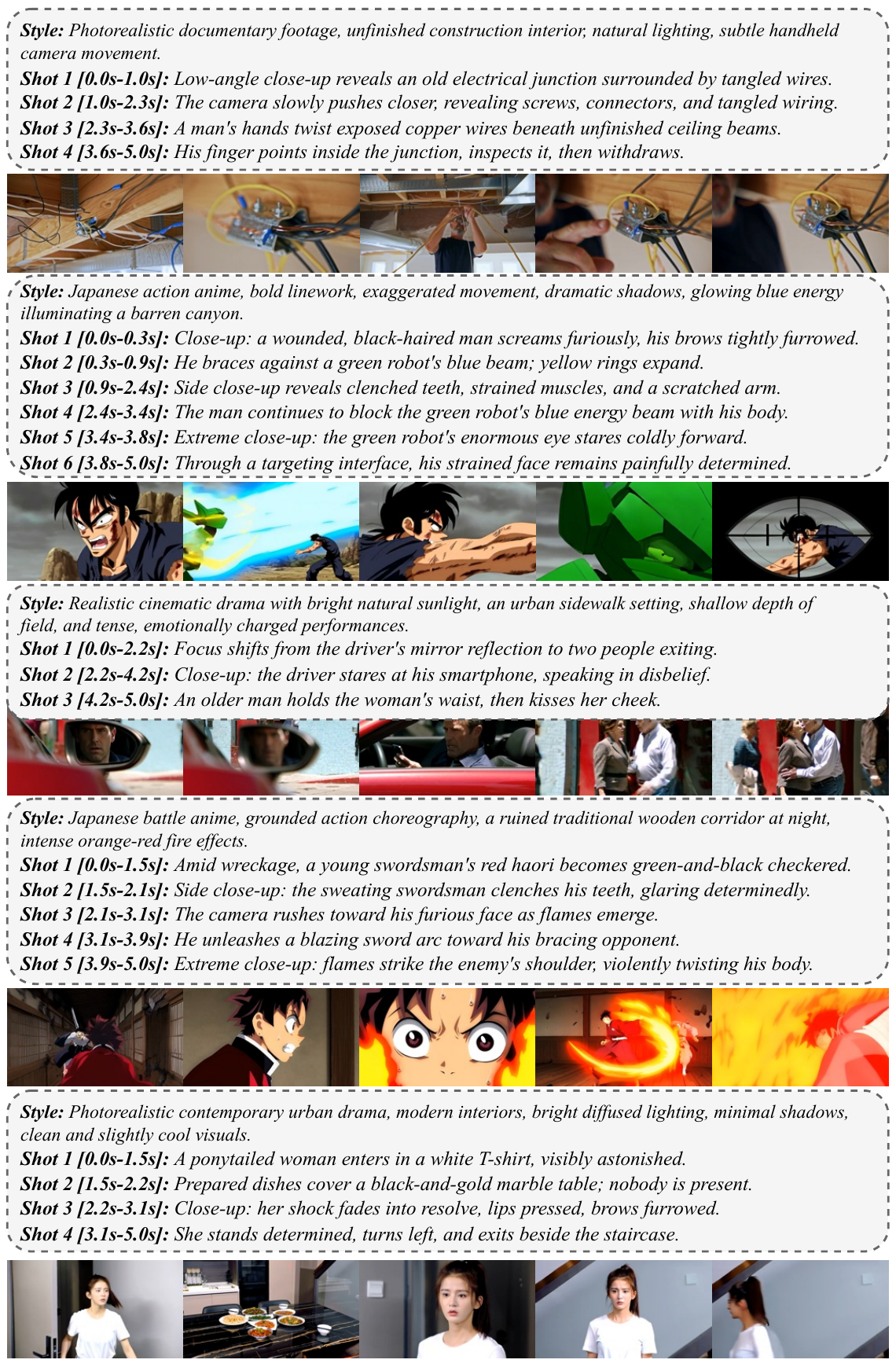}
    \caption{\textbf{Text-only multi-shot generation.} Examples span documentary, animation,
    and drama. Each request specifies a shared style and per-shot events.}
    \label{fig:t2v-supple}
\end{figure}

\begin{figure}[htbp]
    \centering
    \includegraphics[width=0.90\linewidth]{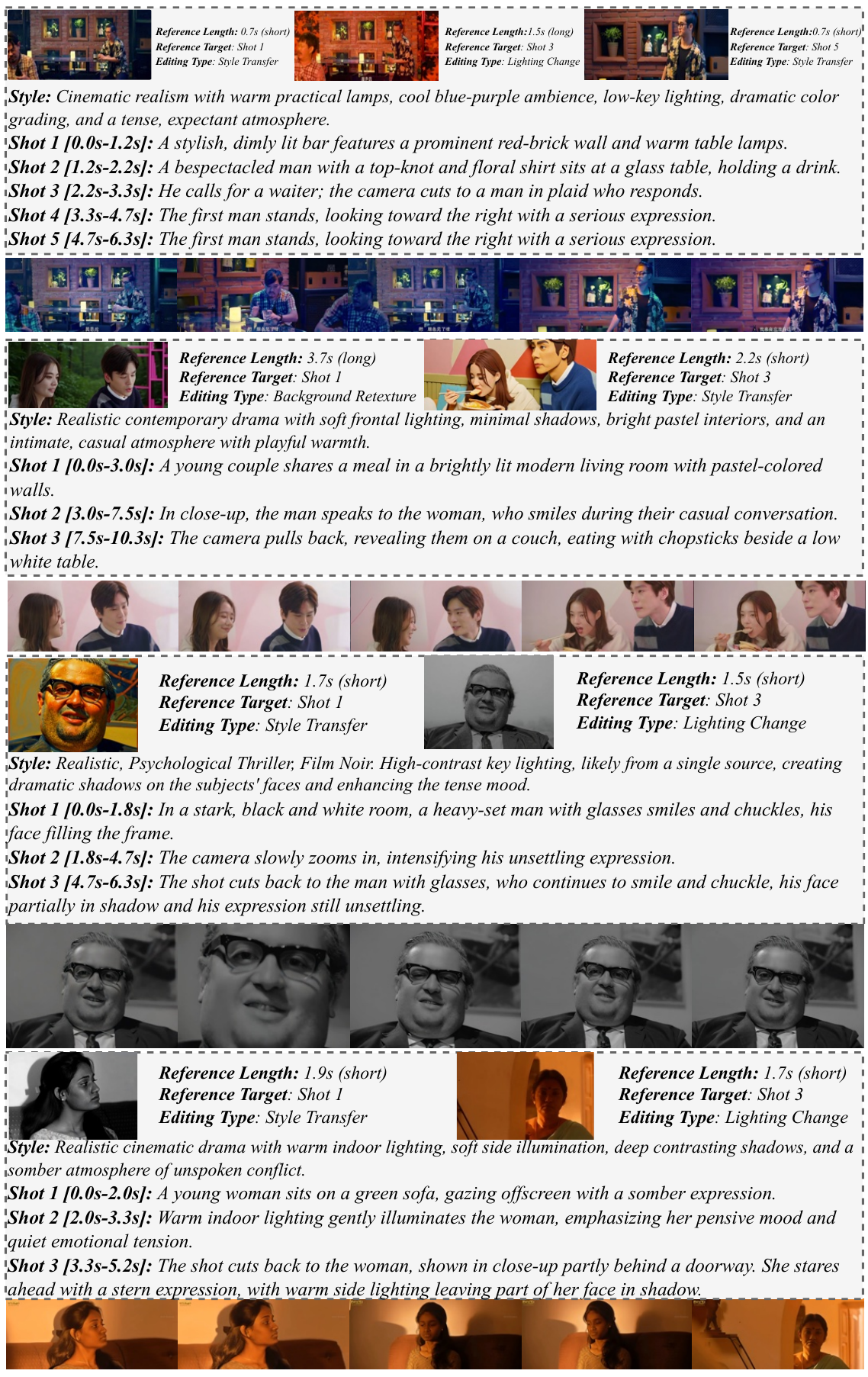}
    \caption{\textbf{Reference-conditioned generation (1/2).} Each example shows independently
    edited references, the requested shared style, and generated shots.}
    \label{fig:r2v-supple-1}
\end{figure}

\begin{figure}[htbp]
    \centering
    \includegraphics[width=0.90\linewidth]{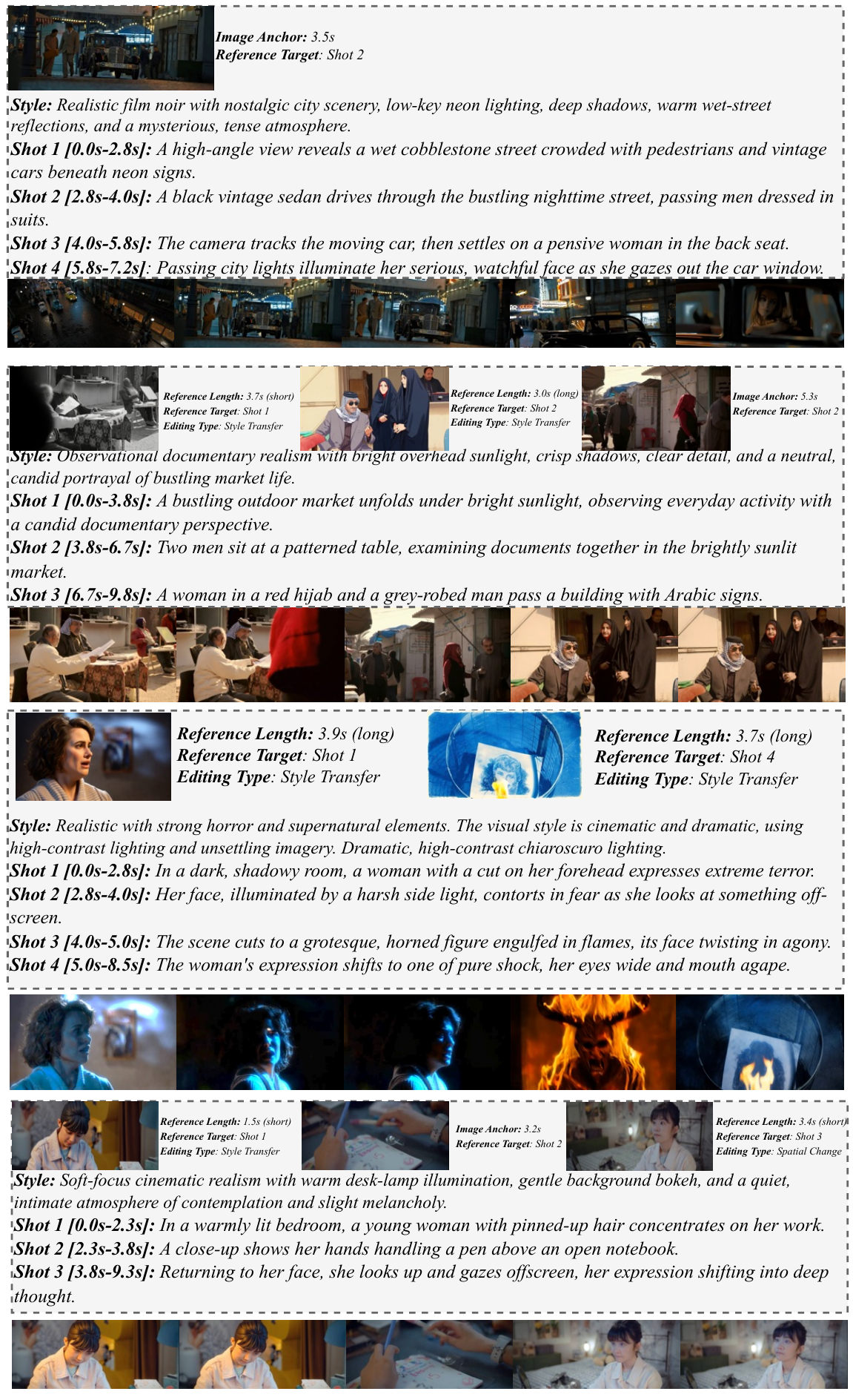}
    \caption{\textbf{Reference-conditioned generation (2/2).} Further examples under the protocol
    of Figure~\ref{fig:r2v-supple-1}.}
    \label{fig:r2v-supple-2}
\end{figure}

\end{document}